\documentclass[11pt]{article}
\pdfoutput=1

\usepackage{mathrsfs}
\usepackage{amsmath,amssymb}
\usepackage{bm}
\usepackage{natbib}
\usepackage[usenames]{color}
\usepackage{amsthm}

\usepackage{multirow} 
\usepackage{enumitem}
\usepackage{caption}
\usepackage{subcaption}
\usepackage{enumitem}
\usepackage{dsfont}
\usepackage{mathtools}
\usepackage{caption}
\usepackage{booktabs}

\usepackage[colorlinks,
linkcolor=red,
anchorcolor=blue,
citecolor=blue
]{hyperref}

\usepackage{mylatexstyle}
\usepackage{float}  
\usepackage{setspace}
\usepackage[left=1in, right=1in, top=1in, bottom=1in]{geometry}

\setlist[itemize]{itemsep=0pt, topsep=2pt}
\setlist[enumerate]{itemsep=-2pt, topsep=2pt}

\usepackage{tcolorbox}
\usepackage{booktabs}
\usepackage{multirow}
\usepackage{threeparttable}
\tcbuselibrary{breakable}
\usepackage{algorithm}
\usepackage{algorithmic}
\usepackage{xcolor}

\ifdefined\final
\usepackage[disable]{todonotes}
\else
\usepackage[textsize=tiny]{todonotes}
\fi
\allowdisplaybreaks

\title{Beyond Consistency: Inference for the Relative risk functional in Deep Nonparametric Cox Models}

\newcommand{\Prob}{\mathbb{P}}

\newcommand{\SE}{\mathrm{SE}}
\newcommand{\EmpSD}{\mathrm{EmpSD}}

\newcommand{\AIL}{\mathrm{AIL}}

\newtheorem{theorem}{Theorem}

\newtheorem{cy}{Corollary}%[section]

\newtheorem{as}{Condition}
\newtheorem{ex}{Example}

\author
{
Sattwik Ghosal \thanks{Department of Biostatistics, University of Michigan; e-mail: {\tt ghosal@umich.edu}}~~~~
Xuran Meng\thanks{Department of Biostatistics, University of Michigan; e-mail: {\tt xuranm@umich.edu}}
	~~~and~~~
Yi Li\thanks{Department of Biostatistics, University of Michigan;
  e-mail: {\tt yili@umich.edu}}
}

\date{}

\begin{document}

\maketitle

\begin{abstract}
There remain theoretical gaps in deep neural network estimators for the nonparametric Cox proportional hazards model. In particular, it is unclear how gradient-based optimization error propagates to population risk under partial likelihood, how pointwise bias can be controlled to permit valid inference, and how ensemble-based uncertainty quantification behaves under realistic variance decay regimes. 
We develop an asymptotic distribution theory for deep Cox estimators that addresses these issues. First, we establish nonasymptotic oracle inequalities for general trained networks that link in-sample optimization error to population risk without requiring the {exact} empirical risk {optimizer}. We then construct a structured neural parameterization that achieves
infinity-norm approximation rates compatible with the oracle bound,
yielding control of the pointwise bias. 
Under these conditions and using the H\'{a}jek--Hoeffding projection, we prove pointwise and multivariate asymptotic normality for subsampled ensemble estimators. We derive a range of subsample sizes that balances bias correction with the requirement that the H\'{a}jek--Hoeffding projection remain dominant. This range accommodates decay conditions on the single-overlap covariance, which measures how strongly a single shared observation influences the estimator, and is weaker than those imposed in the subsampling literature. 
An infinitesimal jackknife representation provides analytic covariance estimation and valid Wald-type inference for relative risk contrasts such as log-hazard ratios. Finally, we illustrate the finite-sample implications of the theory through simulations and a real data application. 
\end{abstract}

\section{Introduction}

\label{sec:intro}
High-complexity estimators based on deep neural networks (DNNs) have achieved success across a range of statistical tasks. Although  progress has been made in establishing population risk consistency and approximation guarantees in some settings \citep{SchmidtHieber2020, FarrellLiangMisra2021},   statistical inference remains underdeveloped. In particular, it is not well understood how optimization error arising from gradient-based training is carried into population risk \citep{Allen-Zhu-AZYL, Arora2019FineGrainedAO}, 
how to enforce the pointwise bias control required for valid confidence intervals \citep{Hall1992, CattaneoFarrellFeng2020, FarrellLiangMisra2021}, and how to construct tractable and reliable uncertainty quantification for scientifically relevant targets such as relative-risk contrasts  \citep{lakshminarayanan2017simple,Chernozhukov2018, WagerAthey2018}.  Moreover, the inferential theory for DNN estimators in the presence of censoring is limited.

We introduce the 
nonparametric Cox proportional hazards model \citep{Cox1972} under independent censoring. Let $\mathbf{X}\in[0,1]^{p_0}$   denote a vector of (normalized) baseline covariates with dimension $p_0$ and  $T_U$ denote the event time. The nonparametric Cox model \citep{Sleeper1990-wi, OSullivan1993-ug, Chen2007-hn, Chen2010-sd} specifies the conditional hazard of $T_U$ given $\mathbf{X}$ as
\begin{align}
\lambda(t \mid \mathbf{X}) 
= \lambda_0(t)\exp\{g_0(\mathbf{X})\},  
\label{eq:Cox model}
\end{align}
where $\lambda_0(\cdot)$ is an unspecified baseline hazard function and $g_0:[0,1]^{p_0}\to\mathbb{R}$ is an unknown  function with  $g_0(\mathbf{0})=0$ for identifiability.  We estimate $g_0$ by $\hat g$ over a DNN function class and also consider relative risk comparisons. 
For two covariate vectors $\mathbf{x}^{(1)}_*$ and $\mathbf{x}^{(2)}_*$, the log-hazard contrast under model~\eqref{eq:Cox model} is
\begin{equation} \label{eq:contrastIntro}
\log\frac{\lambda\big(t \mid \mathbf{X}=\mathbf{x}^{(1)}_*\big)}
         {\lambda\big(t \mid \mathbf{X}=\mathbf{x}^{(2)}_*\big)}
:= \psi\big(\mathbf{x}^{(1)}_*, \mathbf{x}^{(2)}_*\big)
= g_0\big(\mathbf{x}^{(1)}_*\big) - g_0\big(\mathbf{x}^{(2)}_*\big),
\end{equation}
which provides an interpretable measure of relative risk between the two covariate settings.

  Recent semiparametric theory focuses on finite dimensional parameters, 
treating $g_0$ as a machine learned nuisance 
\citep{Chernozhukov2018}. In contrast, inference for $g_0$ itself, particularly for 
DNN estimators, remains largely unresolved. Existing methods establish risk consistency 
but do not control pointwise bias sufficiently to obtain distributional approximations 
for high-complexity DNN estimators 
\citep{Zhong2022-um, Zhou2023-ij, Sun2024-aw, Wu2024-xw}.

We develop inference for DNN estimators with censored survival data 
\citep{Katzman2018DeepSurv, Wiegrebe2024-js}. This setting poses several challenges. 
First, censoring leads to incomplete observations and complicates 
the objective function. Second, practical training algorithms yield only 
approximate empirical risk optimizers, and  it remains unclear how the resulting optimization gap carries over to the population risk in Cox models
\citep{SchmidtHieber2020, Kohler2021-hw, Zhong2022-um, Meng2025-zh}. 
Third, the nonparametric component $g_0(\cdot)$ is estimable at rates slower than $n^{-1/2}$, 
so classical semiparametric theory 
\citep{AndersenGill1982, BickelKlaassenRitovWellner1993, vanderVaart1998} 
does not provide distributional approximations. 
Valid inference requires linking optimization error to population risk, 
controlling pointwise bias, and characterizing the covariance of nonlinear functionals. 
In particular, inference for the contrast 
$\psi\big(\mathbf{x}^{(1)}_*,\mathbf{x}^{(2)}_*\big)$ 
depends on the joint distribution of 
$\hat g(\mathbf{x}^{(1)}_*)$ and $\hat g(\mathbf{x}^{(2)}_*)$. 
Their dependence, induced by the shared training sample, must be accounted for. 
We develop a three-part inferential framework to address these challenges.

% Because DNN estimators permit highly 
% nonlinear representations, valid uncertainty quantification for such 
% relative-risk functionals is central to personalized risk assessment 
% \citep{Zhong2022-um}. However, conducting inference for 
% $\psi(\mathbf{x}^{(1)}_*, \mathbf{x}^{(2)}_*)$ at fixed out-of-sample covariate 
% pairs . Existing 
% theoretical results for deep Cox estimators do not provide such joint 
% distributional guarantees \citep{Zhong2022-um, Sun2024-aw}. 

\begin{enumerate}[leftmargin=*]
 
 \item{\bf Optimization-to-population risk bridge under partial likelihood.} We establish an oracle inequality for DNN estimators trained via gradient-based optimization under the Cox partial likelihood. Existing analyses typically assume access to the exact empirical optimizer \citep{Zhong2022-um}, which permits a two-step argument: first, uniform convergence establishes consistency, and second, a localized empirical process analysis around the true function yields convergence rates.  In contrast, gradient-based training does not guarantee a global optimizer of the empirical objective, preventing a direct application of such localization arguments. Instead, we derive a global bound on the population risk of the form
\[
\text{population risk} \;\lesssim\;
\text{optimization gap}
+ \text{approximation error}
+ \text{statistical error},
\]
where the optimization gap captures the deviation from exact empirical optimality. The resulting bound is self-referential, in that the population risk appears on both sides of the inequality. We resolve this by combining empirical process techniques for Cox models \citep{Huang1999-ir, Zhong2022-um} with Bernstein-type concentration inequalities over suitable coverings of DNN class.
 
 \item{\bf Bias calibration linking population risk and pointwise inference.}
We control pointwise bias via sharp $L_\infty$ approximation bounds. 
By scaling the network architecture (e.g., width and depth), the approximation error is 
asymptotically dominated by the stochastic fluctuation of the estimator, analogous to 
undersmoothing in classical nonparametric inference. This renders the approximation term 
in the oracle inequality negligible for pointwise inference. We further impose a mild 
alignment condition requiring that the trained network achieve the approximation rate, avoiding underutilization of model capacity during optimization.

 \item{\bf Subsample ensemble inference under relaxed single-overlap covariance decay.}
We construct a subsampling-based ensemble estimator and analyze its distribution 
using the Hájek Hoeffding projection. Its variance is characterized by the 
{\em single-overlap covariance} introduced in Section~\ref{sec_theory}, which 
captures dependence between estimators trained on subsamples sharing one observation 
\citep{WagerAthey2018}. Existing ESM theory for generalized nonparametric regression 
\citep{Meng2025-zh} typically assumes near minimal decay of this covariance at rate $1/r$, 
where $r$ is the subsample size. We relax this requirement and establish asymptotic normality 
over a broader range of decay rates. 
A key condition is that $r$ is large enough so that approximation bias is negligible while remaining 
$o(n)$, so that higher-order U-statistic terms are negligible and the Hájek projection dominates. 
Under these conditions, we establish multivariate asymptotic normality for finite sets of evaluation 
points and prove consistency of an infinitesimal jackknife covariance estimator, enabling Wald-type 
inference for nonlinear contrasts \eqref{eq:contrastIntro}.
\end{enumerate}

 \par  Section~\ref{sec:2} reviews DNNs and the nonparametric Cox model. 
Section~\ref{sec3} introduces the subsample-based ensemble inferential framework. 
Section~\ref{sec_theory} presents the theoretical results, including global 
consistency and asymptotic distributional theory for the ensemble estimator. 
Sections~\ref{sec:4} and \ref{sec:realdata} illustrate the finite-sample performance 
through simulations and a real data application. Section~\ref{sec:conc} concludes. 
Full proofs and 
auxiliary lemmas are given in the appendix.  

\subsection{Notation} 
% \ylcm{check AOS. Do they put notation at the end of Section 1 or in the beginning of Section 2?} 
 Vectors are boldfaced: uppercase bold letters (e.g., $\mathbf{X}$) represent random vectors, and lowercase bold letters (e.g., $\mathbf{x}$ or $\mathbf{x}_*$) denote fixed points or realizations.  For random  $\mathbf{X}$, $\Prob_{\mathbf{X}}$ denotes its distribution and $\mathbb{E}_{\mathbf{X}}$ the corresponding expectation.
For a vector $\mathbf{x}=(x_1,\ldots,x_d)^\top$, define its $\ell_p$-, $\ell_\infty$-, and $\ell_0$-norms by
\(
|\mathbf{x}|_p = \Big(\sum_{j=1}^d |x_j|^p\Big)^{1/p}, \qquad
|\mathbf{x}|_\infty = \max_{1\le j\le d}|x_j|, \qquad
|\mathbf{x}|_0 = \sum_{j=1}^d \mathbb{I}(x_j\neq 0),
\)
where $\mathbb{I}(\cdot)$ is the indicator function; for a function $f:\mathcal{X}\to\mathbb{R}$, define its $L_p$ norm by
\(\|f\|_p = \Big(\int_{\mathcal{X}} |f(x)|^p\,dx\Big)^{1/p},  1\le p<\infty,  \,\,
\|f\|_\infty = \sup_{x\in\mathcal{X}} |f(x)|.
\)
For a matrix $\mathbf{A}=(a_{ij})$, define
\(
\|\mathbf{A}\|_0 = \sum_{i,j}\mathbb{I}(a_{ij}\neq 0), \qquad
\|\mathbf{A}\|_\infty = \max_i \sum_j |a_{ij}|.
\)
For two deterministic sequences $x_n$ and $a_n$, write $x_n = O(a_n)$ if $|x_n/a_n|\le C$ for a $C>0$ when $n$ is sufficiently large, and $x_n=o(a_n)$ if $x_n/a_n\to 0$. For random variables $X_n$, $X_n=O_p(a_n)$ if for any $\epsilon>0$ there exists $C>0$ such that $P(|X_n/a_n|>C)\le \epsilon$ when $n$ is sufficiently large, and $X_n=o_p(a_n)$ if $X_n/a_n \overset{p}{\to}0$.
For sequences $X_n$ and $Y_n$,  write $X_n\lesssim Y_n$ if $X_n\le C Y_n$ for a  $C>0$ when $n$ is sufficiently large, and $X_n\asymp Y_n$ if  $X_n\lesssim Y_n$ and $Y_n\lesssim X_n$. Finally, $a\wedge b=\min\{a,b\}$ and $a\vee b=\max\{a,b\}$.
 
%%%%%%%%
\section{Preliminaries}\label{sec:2}

\subsection{Multilayer Neural Networks}

 Let $L \ge 1$ denote the number of hidden layers and 
$\mathbf{p}=(p_0,\ldots,p_{L+1})$ the corresponding layer widths. 
With ReLU activation $\sigma(x)=\max\{x,0\}$ applied componentwise, define 
the shifted activation $\sigma_{\mathbf{v}}(\mathbf{x})=\sigma(\mathbf{x}-\mathbf{v})$. 
We focus on ReLU because of its strong approximation properties and favorable 
optimization behavior; in particular, its derivative equals one for positive 
inputs, preventing gradient attenuation during training 
\citep{SchmidtHieber2020, Zhong2022-um}.

A network with architecture $(L,\mathbf{p})$ is the mapping 
$f:\mathbb{R}^{p_0}\to\mathbb{R}^{p_{L+1}}$ defined by
\begin{align}
f(\mathbf{x})
= \mathcal{W}_L \, \sigma_{\mathbf{v}_L}\!\Big(
\mathcal{W}_{L-1} \, \sigma_{\mathbf{v}_{L-1}}\!\Big(
\cdots 
\mathcal{W}_1 \sigma_{\mathbf{v}_1}\!\big(\mathcal{W}_0 \mathbf{x}\big)
\cdots \Big)\Big),
\label{eq:DNN}
\end{align}
where $\mathcal{W}_l\in\mathbb{R}^{p_{l+1}\times p_l}$ and 
$\mathbf{v}_l\in\mathbb{R}^{p_{l+1}}$. 
Here $p_0$ denotes the input dimension, $p_1,\ldots,p_L$ are the widths 
of the $L$ hidden layers, and $p_{L+1}$ is the output dimension; 
for example, in Figure~\ref{fig:1}, $L=3$ with 
$p_0=5$, $p_1=p_2=4$, $p_3=3$, and $p_4=1$.

To mitigate overfitting, we restrict attention to sparsely connected networks. 
Define the $s$-sparse network class with envelope $F$ by
\begin{align}
\mathcal{F}(L,\mathbf{p},s,F)
=\Big\{ g\in \mathcal{F}(L,\mathbf{p}) :
\sum_{l=0}^{L}\|\mathcal{W}_l\|_{0}
+ \sum_{l=1}^{L}\|\mathbf{v}_l\|_{0} \le s,\ 
\|g\|_{\infty}\le F \Big\},
\label{eq:DNNclass}
\end{align}
where
\begin{align*}
\mathcal{F}(L,\mathbf{p})
=&\Big\{ g(\mathbf{x})=f(\mathbf{x})-f(\mathbf{0}) :
f \text{ is a DNN of the form \eqref{eq:DNN}},\\
&\|\mathcal{W}_l\|_{\infty}\le 1 \text{ for } l=0,\ldots,L,\ 
\|\mathbf{v}_l\|_{\infty}\le 1 \text{ for } l=1,\ldots,L \Big\}.
\end{align*}
By construction, $g(\mathbf{0})=0$ for all $g\in\mathcal{F}(L,\mathbf{p})$, 
which ensures identifiability. 
\begin{center}
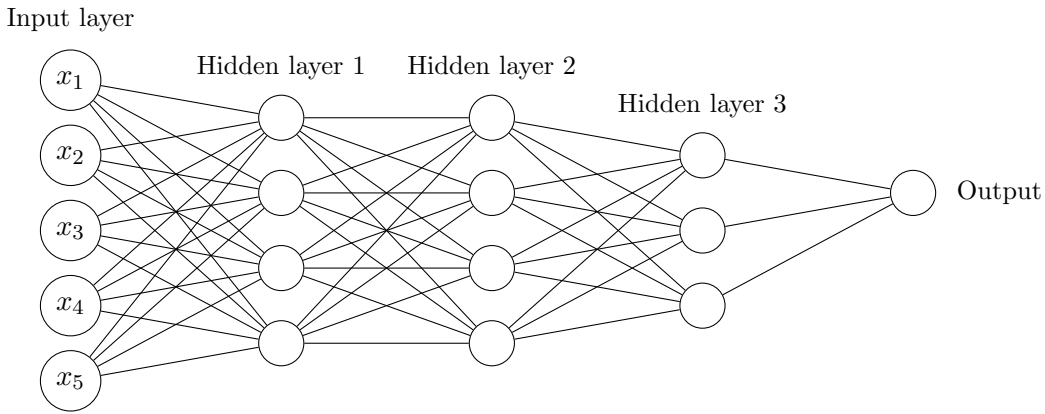
\begin{figure}[htbp]
\begin{tikzpicture}[
  x=1.4cm, y=1cm,
  neuron/.style={circle, draw, minimum size=6mm},
  layerlabel/.style={font=\small}
]

%--------------------
% Input layer: p0 = 5
%--------------------
\foreach \i in {1,...,5} {
  \node[neuron] (I-\i) at (0, -\i) {$x_{\i}$};
}
\node[layerlabel, above=3pt of I-1] {Input layer};

%--------------------
% Hidden layer 1: p1 = 4
%--------------------
\foreach \i in {1,...,4} {
  \node[neuron] (H1-\i) at (2, -0.5-\i) {};
}
\node[layerlabel, above=2pt of H1-1] {Hidden layer 1};

%--------------------
% Hidden layer 2: p2 = 4
%--------------------
\foreach \i in {1,...,4} {
  \node[neuron] (H2-\i) at (4, -0.5-\i) {};
}
\node[layerlabel, above=2pt of H2-1] {Hidden layer 2};

%--------------------
% Hidden layer 3: p3 = 3
%--------------------
\foreach \i [count=\k] in {1,...,3} {
  % shift vertically to roughly center w.r.t others
  \node[neuron] (H3-\i) at (6, -1-\k) {};
}
\node[layerlabel, above=2pt of H3-1] {Hidden layer 3};

%--------------------
% Output layer: p4 = 1
%--------------------
\node[neuron] (O-1) at (8, -2.5) {};
\node[layerlabel, right=4pt of O-1] {Output};

%--------------------
% Connections
%--------------------
% Input -> Hidden 1
\foreach \i in {1,...,5}
  \foreach \j in {1,...,4}
    \draw (I-\i) -- (H1-\j);

% Hidden 1 -> Hidden 2
\foreach \i in {1,...,4}
  \foreach \j in {1,...,4}
    \draw (H1-\i) -- (H2-\j);

% Hidden 2 -> Hidden 3
\foreach \i in {1,...,4}
  \foreach \j in {1,...,3}
    \draw (H2-\i) -- (H3-\j);

% Hidden 3 -> Output
\foreach \i in {1,...,3}
  \draw (H3-\i) -- (O-1);
 %\ylcm{take care of caption. in the text it says Figure 2.1}
\end{tikzpicture}
\caption{A fully connected neural network with $L = 3$ and $\mathbf{p} = (5,4,4,3,1)$.}
  \label{fig:1}
\end{figure}
\end{center}

\subsection{Nonparametric Cox Models and  H\"older Functional Classes}
  %We consider right-censored survival data with event time $T_{i,U}$ and censoring time $T_{i,C}$, for $i=1,\ldots,n$.  We observe $\mathcal D_i=(T_i,\Delta_i,\mathbf X_i)$, %\ylcm{later you used $D_i$ in Section 4}
 % where 
%$T_i=\min(T_{i,U},T_{i,C})$, $\Delta_i=\mathbb{I}(T_{i,U}\le T_{i,C})$, and $\mathbf X_i\in [0,1]^{p_0}$.
%5We assume that $ T_{i,U}$ and $T_{i,C}$ are independent conditional on $\mathbf X_i$ and
 %that $\mathcal D_i$ are independent and identically distributed (i.i.d.). We study the nonparametric Cox model \eqref{eq:Cox model}, which generalizes the classical Cox model \cite{Cox1972-ft} by allowing the log-risk $g_0(\cdot)$ to be an unknown nonlinear function of $\mathbf X$.  
  We consider right-censored survival data, where $T_U$ and $T_C$ denote the 
event time and censoring time, respectively, and 
$\mathbf{X}\in[0,1]^{p_0}$ is a $p_0$-dimensional covariate vector. 
We assume that $T_U$ and $T_C$ are conditionally independent given $\mathbf{X}$ 
\citep{Fleming2005-lk}. 
We observe $n$ independent and identically distributed (i.i.d.)\ copies $D_i=(T_i,\Delta_i,\mathbf{X}_i)$ of 
$D=(T,\Delta,\mathbf{X})$, where 
$T_i=T_{i,U}\wedge T_{i,C}$ and 
$\Delta_i=\mathbb{I}(T_{i,U}\le T_{i,C})$. 
Let $\mathcal{D}_n=\{D_i\}_{i=1}^n$ denote the sample of size $n$. We study the nonparametric Cox model \eqref{eq:Cox model}, which extends 
the classical Cox model \citep{Cox1972} by allowing the log-risk function 
$g_0(\cdot)$ to be an unknown function of $\mathbf{X}$. 
To ensure identifiability, we impose $g_0(\mathbf{0})=0$. 
While existing work on nonparametric and DNN-based estimators mainly establishes 
consistency \citep{Zhong2022-um}, we develop inference for $g_0(\cdot)$, including 
uncertainty quantification for pointwise evaluations and contrasts.

 We assume that $g_0$ belongs to a H\"older class 
$\mathcal{G}_{p_0}^{\gamma}([0,1]^{p_0},K)$ with smoothness $\gamma>0$ 
and radius $K>0$ \citep{SchmidtHieber2020, Zhong2022-um}, defined as
\begin{align*}
\mathcal{G}_{p_0}^{\gamma}([0,1]^{p_0},K)
=
\Bigg\{
g:[0,1]^{p_0}\to\mathbb{R}:\;
&\sum_{\|\beta\|_1<\gamma}\|\partial^{\beta}g\|_\infty \\
&+\!\!\sum_{\|\beta\|_1=\lfloor\gamma\rfloor}
\sup_{\mathbf{x}\ne\mathbf{y}\in[0,1]^{p_0}}
\frac{|\partial^{\beta}g(\mathbf{x})-\partial^{\beta}g(\mathbf{y})|}
{\|\mathbf{x}-\mathbf{y}\|_\infty^{\gamma-\lfloor\gamma\rfloor}}
\le K
\Bigg\}.
\end{align*}
Here $\lfloor\gamma\rfloor$ denotes the integer part of $\gamma$, and for a multi-index 
$\boldsymbol{\beta}=(\beta_1,\ldots,\beta_{p_0})^\top\in\mathbb{N}^{p_0}$, 
$\partial^{\beta}=\partial^{\beta_1}\cdots\partial^{\beta_{p_0}}$.

 We further assume that $g_0$ admits a compositional representation with 
H\"older-smooth components. For some integer $q\ge 0$ and vectors 
$\mathbf{d}=(d_0,\ldots,d_{q+1})\in\mathbb{N}_+^{q+2}$, 
$\mathbf{t}=(t_0,\ldots,t_q)\in\mathbb{N}_+^{q+1}$, and 
$\boldsymbol{\gamma}=(\gamma_0,\ldots,\gamma_q)\in\mathbb{R}_+^{q+1}$, 
we assume
\begin{align}
g_0 \in \mathcal{G}(q,\mathbf{d},\mathbf{t},\boldsymbol{\gamma},K)
=\Big\{ g=g^{(q)}\circ\cdots\circ g^{(0)} : 
& g^{(i)}=(g^{(i)}_{j})_j:[a_i,b_i]^{d_i}\to[a_{i+1},b_{i+1}]^{d_{i+1}}, \nonumber\\
& g^{(i)}_{j}\in\mathcal{G}_{t_i}^{\gamma_i}([a_i,b_i]^{t_i}),\quad |a_i|,|b_i|\le K
\Big\}.
\label{eq:true_functional_class}
\end{align}

   The class $\mathcal{G}$ combines H\"older smoothness with a compositional structure, 
with each $g^{(i)}$ acting on the output of the preceding map, reflecting the layered 
structure of DNNs \citep{Yarotsky2017-hs}. 
The vectors $\mathbf d$ and $\mathbf t$ specify layer-wise dimensions, and 
$\boldsymbol{\gamma}$ encodes smoothness at each level. 
In a $p_0$-variate Cox model \eqref{eq:Cox model}, $g_0:[0,1]^{p_0}\to\mathbb{R}$, 
so $d_0=p_0$, $d_{q+1}=1, a_0=0$ and $b_0=1$. 
We provide two examples with $q=2$.
\begin{ex}
\begin{align*}
g_0(\mathbf{x})
=g^{(2)}_{1}\!\Big(
g^{(1)}_{1}\big(g^{(0)}_{1}(x_1,x_2),\,g^{(0)}_{2}(x_3,x_4,x_5)\big),
g^{(1)}_{2}\big(g^{(0)}_{3}(x_6,x_7),\,g^{(0)}_{4}(x_8,x_9,x_{10})\big)
\Big),~ \mathbf{x}\in[0,1]^{10}.
\end{align*}
If all component functions $g^{(i)}_{j}$ are twice continuously differentiable, then  
$\boldsymbol{\gamma}=(2,2,2)$, $\mathbf{d}=(10,4,2,1)^\top$, and $\mathbf{t}=(3,2,2)^\top$.\end{ex}
\begin{ex}
\[
g_0(\mathbf{x})
=g^{(2)}_{1}\!\Big(
g^{(1)}_{1}(g^{(0)}_{1}(x_1,x_2),\,g^{(0)}_{2}(x_3)),
g^{(1)}_{2}(g^{(0)}_{3}(x_4,x_5),\,g^{(0)}_{4}(x_6))
\Big),
\hspace{2mm} \mathbf{x}\in[0,1]^6,
\]
where we suppose each $g^{(i)}_{j}$ has Hölder smoothness $\gamma_i=\frac{3}{2}$.  
Hence, $\boldsymbol{\gamma}=(\frac{3}{2}, \frac{3}{2}, \frac{3}{2})$, $\mathbf{d}=(6,4,2,1)^\top$, and $\mathbf{t}=(2,2,2)^\top$.
\end{ex}
 The composite-smoothness class $\mathcal{G}(q,\mathbf d,\mathbf t,\boldsymbol{\gamma},K)$ 
subsumes standard nonparametric models, including additive structures \citep{Stone1985-pp} 
and nested compositional regressions \citep{Horowitz2007-ct}. Related work extends this 
framework to broader hierarchical compositions beyond the classical H\"older setting 
\citep{Kohler2021-hw}. 

 Define the effective smoothness at level $i$ by
\[
\gamma_i^{*}=\gamma_i\prod_{l=i+1}^{q}(\gamma_l\wedge1),
\]
which represents the H\"older smoothness associated with level $i$ after attenuation through subsequent compositions. With a training sample of size $n$, define
\begin{align}
\phi_n=\max_{i=0,\ldots,q}n^{-2\gamma_i^{*}/(2\gamma_i^{*}+t_i)},
\end{align}
which is the minimax-optimal estimation rate, up to logarithmic factors, for deep ReLU estimators  \citep{SchmidtHieber2020}. Writing $\phi_n=n^{-c_{\mathrm{eff}}}$ gives
\begin{equation}
\label{junk}
c_{\mathrm{eff}}=\min_{i=0,\ldots,q}\frac{2\gamma_i^{*}}{2\gamma_i^{*}+t_i},
\qquad
i_{\mathrm{eff}}=\arg\min_{0\le i\le q}\frac{2\gamma_i^{*}}{2\gamma_i^{*}+t_i}.
\end{equation}
Here, $i_{\mathrm{eff}}$ identifies the bottleneck level in the compositional structure, and $c_{\mathrm{eff}}$ determines the overall rate $\phi_n=n^{-c_{\mathrm{eff}}}$. Thus, the slowest level in the H\"older hierarchy governs this overall  rate. It also sets the scale of the optimal approximation bias as stated in  Theorem~\ref{thm:approx_error}.

\section{Inference on DNN Estimates via an   Ensemble Subsampling Learner} \label{sec3} 
 We estimate \( g_0(\cdot) \) in model~\eqref{eq:Cox model} by maximizing the  Cox partial likelihood over the DNN function class \( \mathcal{F}(L,s,\mathbf{p},F) \) defined in~\eqref{eq:DNNclass}. For $g:[0,1]^{p_0}\to\mathbb{R}$, the log partial likelihood is 
\begin{align}
L_n(g)
= n^{-1}\sum_{i=1}^{n}\Delta_i\Big[g(\mathbf{X}_i)-\log\Big\{\sum_{j:\,T_j\ge T_i}\exp\{g(\mathbf{X}_j)\}\Big\}\Big].
\label{eq:partlkhd}
\end{align}
The empirical optimizer is
\begin{equation}
\widehat{g}_{\text{opt}} = \arg\max_{g \in \mathcal{F}(L,\mathbf{p},s,F)}L_n(g).
\label{eq: g_hat DNN}
\end{equation}
In practice, DNNs are trained by stochastic gradient methods and may not attain $\widehat{g}_{\text{opt}}$. To quantify the resulting optimization error, we define the expected optimization gap
\begin{align*}
\mathcal{Q}_n(\hat g, \widehat{g}_{\text{opt}})
:= \mathbb{E}_{\mathcal{D}_n}\{ L_n( \widehat{g}_{\text{opt}}) - L_n(\hat g)\},
\end{align*}
so that $\mathcal{Q}_n(\hat g,\widehat{g}_{\text{opt}}) \ge 0$, with equality if $\hat g = \widehat{g}_{\text{opt}}$. When the context is clear, we write $\mathcal{Q}_n(\hat g):=\mathcal{Q}_n(\hat g,\widehat{g}_{\text{opt}})$.

Unlike separable losses, \eqref{eq:partlkhd} does not decompose into a sum of independent contributions due to risk-set coupling, and standard empirical process arguments do not  apply. To isolate this dependence, we rewrite the partial likelihood as
\begin{equation} 
\label{eq:partlkhd0}
L_n(g)
= n^{-1}\sum_{i=1}^n \Delta_i\,\ell_n(T_i,\mathbf{X}_i;g)  - n^{-1}\sum_{i=1}^n \Delta_i,
\end{equation}
where 
\[
\ell_n(t,\mathbf{X};g)=g(\mathbf{X})-\log S_n^{(0)}(t;g), \qquad
S_n^{(0)}(t;g)=n^{-1}\sum_{j=1}^n Y_j(t)\exp\{g(\mathbf{X}_j)\},
\]
and $Y_j(t)=\mathbb{I}(T_j\ge t)$. Since the second term in \eqref{eq:partlkhd0} does not depend on $g$, we  work with
\[
L_n(g)= n^{-1}\sum_{i=1}^n \Delta_i\,\ell_n(T_i,\mathbf{X}_i;g),
\]
which yields the same optimizer as \eqref{eq: g_hat DNN}.

We define the population counterpart of $L_n(\cdot)$ as
\[
L_{0}(g)=\mathbb{E}_D\big\{\Delta\,\ell_0(T,\mathbf{X};g)\big\},
\qquad
\ell_0(t,\mathbf{X};g)=g(\mathbf{X})-\log S^{(0)}(t;g),
\]
for $0\le t \le \tau$, where $\tau$ is the study end time (Condition~\ref{as:nondegcox}), and
\[
S^{(0)}(t;g)=\mathbb{E}_D\{Y(t)\exp\{g(\mathbf{X})\}\}, \qquad
Y(t)=\mathbb{I}(T\ge t).
\]
The population risk of any fixed function $g$ is
\[
R(g,g_0)=L_0(g_0)-L_0(g),
\]
which is nonnegative (Lemma~1 in appendix). For a data-dependent estimator $\hat g$, the risk $R(\hat g,g_0)$ is random, and we consider its expectation
\[
\mathbb{E}_{\mathcal D_n}[R(\hat g,g_0)],
\]
which we refer to as the population risk of $\hat g$.

Uniform convergence of $S_n^{(0)}(t;g)$ to $S^{(0)}(t;g)$ over $t\in[0,\tau]$ allows the data-dependent risk-set average to be replaced by its deterministic limit, so that $L_n(g)$ can be treated as an empirical process with a remainder term, whose magnitude we quantify for DNN estimators.

Inference based on $\hat g$ is challenging because it lacks a tractable influence-function representation and its sampling variability is difficult to characterize. By averaging DNN estimators trained on subsamples, ESM stabilizes the estimator and yields an approximate linear representation via its first-order H\'ajek projection, enabling statistical inference.
\subsection{Estimation and Inference via Ensemble Subsampling}

\noindent
 For a fixed covariate point $\mathbf{x}_*\in[0,1]^{p_0}$ and a pair 
$\mathbf{x}_*^{(1)},\mathbf{x}_*^{(2)}\in[0,1]^{p_0}$, we introduce an 
ensemble subsampling method (ESM; Algorithm~1) to estimate $g_0(\mathbf{x}_*)$ 
and conduct inference for both $g_0(\mathbf{x}_*)$ and the log-hazard ratio 
$\psi(\mathbf{x}_*^{(1)},\mathbf{x}_*^{(2)}) 
= g_0(\mathbf{x}_*^{(1)}) - g_0(\mathbf{x}_*^{(2)})$.
The ensemble estimator $\hat{g}_{B}$ aggregates subsample estimators 
$\{\widehat g^b\}$ trained on overlapping subsets of $\mathcal D_n$, and 
thus forms a sum of dependent terms. However, $\hat{g}_{B}$ has the structure 
of an incomplete U-statistic over subsamples. This enables a H\'ajek projection 
\citep{WagerAthey2018}, which decomposes $\hat{g}_{B}$ into a first-order term, 
a sum of independent influence contributions, and an asymptotically negligible 
higher-order remainder. The  linear representation characterizes the 
asymptotic distribution.

\begin{tcolorbox}[breakable, colback=white, colframe=black!40!gray, title={Algorithm 1: Ensemble Subsampling}]
\begin{algorithmic}[1]
\small

\STATE \textbf{Input:} Observed  data $\mathcal{D}_n=\{(T_i,\Delta_i,\mathbf{X}_i):i=1,\ldots,n\}$, subsample size $r$, number of subsamples $B$.
\vspace{0.8ex}

\STATE \textbf{Subsampling:} 
 Let $\mathcal{I}=\{1,\ldots,n\}$ index observations. Using subset sampling, randomly generate 
$B$ subsets $\mathcal{I}^b \subset \mathcal{I}$ with $b=1,\ldots,B$.
\vspace{0.8ex}
\STATE \textbf{Model Training:} 

 For each $1 \le b \le B$, we maximize the subsample partial likelihood 
$L_r(g;\mathcal{I}^b)$ defined from \eqref{eq:partlkhd} on the observations indexed by $\mathcal{I}^b$.   
The optimization is performed  over 
$\mathcal{F}(L,\mathbf{p},s,F)$ using a gradient-based algorithm such as SGD.  
Let $\widehat{g}^b$ denote the resulting (approximate) solution,
\begin{equation}
\widehat{g}^b \approx \widehat{g}^{b}_{\mathrm{opt}}
:= \arg\max_{g \in \mathcal{F}(L,\mathbf{p},s,F)}
L_r(g;\mathcal{I}^b),
\label{eq:g_hat_SGD}
\end{equation}
 so that
\begin{equation}
\mathcal{Q}_r(\widehat{g}^b,\widehat{g}^{b}_{\mathrm{opt}})
\le \mathcal{Q}_b^{\mathrm{opt}},
\label{eq:SGC-opt}
\end{equation}
where  $\mathcal{Q}_b^{\mathrm{opt}}$ quantifies the optimization error.

\vspace{0.8ex}
\STATE \textbf{Ensemble Estimator:}
 Aggregate the subsample estimators to form the ensemble estimate for any fixed   $\mathbf{x}_*\in[0,1]^{p_0}$  
\begin{align}
\hat{g}_{B}(\mathbf{x}_*)=\frac{1}{B}\sum_{b=1}^{B}\widehat{g}^b(\mathbf{x}_*).
\label{eq: ESM g}
\end{align}
%Since each $\widehat{g}^b(\mathbf{0})=0$, the ensemble also satisfies $\hat{g}_{B}(\mathbf{0})=0$. 
The corresponding log-hazard ratio (contrast) estimator for two profiles $\mathbf{x}_*^{(1)}$ and $\mathbf{x}_*^{(2)}$ is
\begin{align}
\widehat{\psi}_{B}(\mathbf{x}_*^{(1)},\mathbf{x}_*^{(2)})
=
\hat{g}_{B}(\mathbf{x}_*^{(1)})-
\hat{g}_{B}(\mathbf{x}_*^{(2)}).
\label{bicontra}
\end{align}
\vspace{0.8ex}
\STATE \textbf { Variance Estimation:}
 Estimate the asymptotic pointwise variance $\widehat{\sigma}^2(\mathbf{x}_*)$ and the contrast variance $\widehat{\sigma}^2_{1,2}$ using the infinitesimal jackknife (IJ), which leverages the ESM subsampling structure to provide uncertainty quantification.

\STATE \textbf{Confidence Interval (CI) Construction:}
 Compute the approximate $(1-\alpha) \times 100 \%$ CIs:
\begin{align*}
&\mathrm{CI}\big(g_0(\mathbf{x}_*)\big)
=
\Big[
\widehat{g}_{B}(\mathbf{x}_*) 
- c_\alpha\,\widehat{\sigma}(\mathbf{x}_*), 
\widehat{g}_{B}(\mathbf{x}_*) 
+ c_\alpha\,\widehat{\sigma}(\mathbf{x}_*)
\Big],\\
&\mathrm{CI}\big(\psi(\mathbf{x}^{(1)}_*,\mathbf{x}^{(2)}_*)\big)
=
\Big[
\widehat{\psi}_{B}(\mathbf{x}^{(1)}_*,\mathbf{x}^{(2)}_*) 
- c_\alpha\,\widehat{\sigma}_{1,2}, 
\widehat{\psi}_{B}(\mathbf{x}^{(1)}_*,\mathbf{x}^{(2)}_*) 
+ c_\alpha\,\widehat{\sigma}_{1,2}
\Big],
\end{align*} 
where $ 0<\alpha<1$ and  $c_\alpha$ denotes the $(1-\alpha/2)$ quantile of standard normal.
With $\frac{\lambda(t\mid \mathbf{x}^{(1)}_*)}{\lambda(t\mid \mathbf{x}^{(2)}_*)} = \exp\!\big(\psi(\mathbf{x}^{(1)}_*,\mathbf{x}^{(2)}_*)\big)$, the corresponding CI for  relative risk is  
\(
\big[\exp(L), \exp(U)\big],
\)
where $L$ and $U$ are the lower and upper bounds of $\mathrm{CI}\big(\psi(\mathbf{x}^{(1)}_*,\mathbf{x}^{(2)}_*)\big)$, respectively.

\end{algorithmic}
\end{tcolorbox}

Enumerating all $B^\ast=\binom{n}{r}$ subsets is infeasible, so we draw $B$ random subsets $\mathcal{I}^{b_1},\ldots,\mathcal{I}^{b_B}$    $\subset\mathcal{I}=\{1,\ldots,n\}$ 
with $|\mathcal{I}^b|=r$. The ESM estimator is defined as the average of the 
corresponding subsample estimators in \eqref{eq: ESM g}.
  
\par  We estimate the asymptotic variance of the ensemble estimate $\widehat g_{B}(\mathbf{x}_*)$ using the bias-corrected infinitesimal jackknife (IJ) variance estimator, denoted as $\hat{\sigma}^2(\mathbf{x}_*)$. Define 
\[
Z_{ji}(\mathbf{x}_*)
=
\big(J_{ji}-J_{\cdot i}\big)\big(\widehat g^{b_j}(\mathbf{x}_*)-\widehat g_{B}(\mathbf{x}_*)\big),
\qquad
\widehat V_i(\mathbf{x}_*)
=
\frac{1}{B}\sum_{j=1}^B Z_{ji}(\mathbf{x}_*),
\]
where $J_{ji}=I(i\in\mathcal I^{b_j})$ is the subsample inclusion indicator for replicate $j$, and $J_{\cdot i}=B^{-1}\sum_{j=1}^B J_{ji}$. The pointwise IJ variance is then given by
\begin{align}
\hat{\sigma}^2(\mathbf{x}_*)
=&
\frac{n(n-1)}{(n-r)^2}
\bigg\{
\sum_{i=1}^n \widehat V_i^2(\mathbf{x}_*)
-
\frac{1}{B(B-1)}
\sum_{i=1}^n\sum_{j=1}^B
\big(Z_{ji}(\mathbf{x}_*)-\widehat V_i(\mathbf{x}_*)\big)^2
\bigg\}.
\label{eq:sigma_pointwise}
\end{align}

Consequently, to evaluate the relative risk between two covariate profiles, we estimate the variance of the contrast $\widehat{\psi}_{B}(\mathbf{x}^{(1)}_*, \mathbf{x}^{(2)}_*) = \widehat g_{B}(\mathbf{x}^{(1)}_*)-\widehat g_{B}(\mathbf{x}^{(2)}_*)$ as
\begin{equation}
\hat{\sigma}^2_{1,2}
=
\hat{\sigma}^2(\mathbf{x}^{(1)}_*)+\hat{\sigma}^2(\mathbf{x}^{(2)}_*)-2\hat{\tau}(\mathbf{x}^{(1)}_*,\mathbf{x}^{(2)}_*),
\label{variance est}
\end{equation}
where 
\begin{align}
\hat{\tau}(\mathbf{x}^{(1)}_*,\mathbf{x}^{(2)}_*)
=&
\frac{n(n-1)}{(n-r)^2}
\bigg\{
\sum_{i=1}^n \widehat V_i(\mathbf{x}^{(1)}_*)\,\widehat V_i(\mathbf{x}^{(2)}_*)  \nonumber \\
 & -
\frac{1}{B(B-1)}
\sum_{i=1}^n\sum_{j=1}^B
\big(Z_{ji}(\mathbf{x}^{(1)}_*)-\widehat V_i(\mathbf{x}^{(1)}_*)\big)
\big(Z_{ji}(\mathbf{x}^{(2)}_*)-\widehat V_i(\mathbf{x}^{(2)}_*)\big)
\bigg\}.
\label{eq:tau_12}
\end{align}

 The variance framework in \eqref{eq:sigma_pointwise}-\eqref{eq:tau_12} 
extends \citet{Meng2025-zh} to account for joint dependence across multiple 
evaluations. The scaling factors correct Monte Carlo and finite-sample bias, 
enabling valid inference when $B \gtrsim n$. Consistency of this   covariance 
estimator is established later.

\section{Theoretical Results}\label{sec_theory}
 We establish asymptotic results for the ESM estimators \eqref{eq: ESM g} and \eqref{bicontra}. 
To characterize their dependence structure, we introduce covariance measures that quantify 
the influence of individual observations.

Let $D_i=(T_i,\Delta_i,\mathbf X_i)$, $i=1,\ldots,n$, be  i.i.d.\ observations. 
Define the single-overlap covariance measures
\begin{equation}
\begin{aligned}
\zeta_{r}^{(1)}(\mathbf{x}_*) 
&= \operatorname{Cov}\!\big\{ \widehat{g}(\mathbf{x}_*; D_1,\ldots,D_r),\,
\widehat{g}(\mathbf{x}_*; D_1,D_2',\ldots,D_r') \big\}, \\
\zeta_{r}^{(1)}(\mathbf{x}_*,\mathbf{x}'_*) 
&= \operatorname{Cov}\!\big\{ \widehat{g}(\mathbf{x}_*; D_1,\ldots,D_r),\,
\widehat{g}(\mathbf{x}'_*; D_1,D_2',\ldots,D_r') \big\},
\end{aligned}
\label{eq: xi_1,r}
\end{equation}
where $\mathbf{x}_*,\mathbf{x}'_*\in [0,1]^{p_0}$ are fixed evaluation points, 
and $D_i'$ are independent copies of $D_i$. 
Here $\widehat{g}(\mathbf{x}_*; D_1,\ldots,D_r)$ denotes the estimator trained on 
$\{D_1,\ldots,D_r\}$ and evaluated at $\mathbf{x}_*$.

These quantities measure the covariance between estimators trained on subsamples 
that share one observation and differ in all remaining entries. They capture the 
leading-order dependence induced by overlap in the subsampling procedure and are 
central to the H\'ajek expansion of the resulting U-statistic 
\citep{WagerAthey2018}.
%%%%%%%%Conditions%%%%%%%%%% 

\begin{as}\label{as:nondegcox}
Assume the study ends at time $\tau$ such that there exists $\varepsilon_0 > 0$ satisfying
\[
P(\Delta = 1 \mid \mathbf{X}) \ge \varepsilon_0
\quad \text{and} \quad
P(T_U \ge \tau \mid \mathbf{X}) \ge \varepsilon_0,
\]
almost surely with respect to ${\Prob}_{\mathbf{X}}$; we also impose the condition $g_0(\mathbf{0}) = 0$.
\end{as}
 Condition~\ref{as:nondegcox} imposes a standard nondegeneracy condition 
\citep{AndersenGill1982}, ensuring that the probabilities of failure and of 
remaining at risk up to time $\tau$ are uniformly bounded away from zero, 
thereby preventing degeneration of the risk-set and partial-likelihood processes. 
The constraint $g_0(\mathbf 0)=0$ ensures identifiability of model \eqref{eq:Cox model}.

To characterize the approximation capacity of the network, we impose architectural 
constraints on the DNN class in the following Condition \ref{as:DNN_structures}. In particular, we introduce 
a hyper parameter $\delta \ge 0$. The case $\delta=0$ corresponds to the baseline regime 
achieving the minimax-optimal rate \citep{SchmidtHieber2020, Zhong2022-um}, whereas 
$\delta>0$ yields an overparameterized architecture required for subsequent analysis.
\begin{as}\label{as:DNN_structures}
With $g_0$ specified in \eqref{eq:true_functional_class}, the network class $\mathcal{F}(L,\mathbf{p},s,F)$ is required to satisfy the following structural conditions for some $\delta \geq 0$,
\begin{itemize}
\item $\log_2 n \lesssim L \lesssim \log^{\mu}(n),$ \quad for some $\mu>1$, 
\item $n^{1+\delta}\phi_n \lesssim \min_{1\le i\le L} p_i \le \max_{1\le i\le L} p_i \lesssim n,$
\item $s \asymp n^{1+\delta}\phi_n\log n,$
\item $F \ge \max\{K,1\}.$
\end{itemize}
\end{as}
   We define the {\em optimal network} in $\mathcal{F}(L,\mathbf{p},s,F)$ that best 
approximates $g_0$:
\[
g_{\textrm{apx}} \in \arg\min_{g \in \mathcal{F}(L,\mathbf{p},s,F)} 
\|g - g_0\|_\infty 
\]
and use the optimal approximation error, $\|g_{\textrm{apx}} - g_0\|_\infty$,  to measure the approximation 
capability of $\mathcal{F}(L,\mathbf{p},s,F)$.  We establish a bound for  $\|g_{\textrm{apx}}-g_0\|_\infty$ and show  that increasing $\delta$ sharpens this bound.

\begin{theorem}[Bound of Optimal Approximation Error]\label{thm:approx_error} Let the true function $g_0$ be specified in \eqref{eq:true_functional_class} with $g_0(\mathbf{0}) = 0$. Under Condition \ref{as:DNN_structures},  it holds that \[ \|g_{\textrm{apx}}-g_0\|^2_{\infty} \lesssim \phi_n^{1+\eta}, \] where $\eta = \frac{2\delta\gamma_{i_{\text{eff}}}^*}{t_{i_{\text{eff}}}c_{\text{eff}}}$, with $c_{\text{eff}}$   and $i_{\text{eff}}$ defined in \eqref{junk}. \end{theorem}

  Theorem~\ref{thm:approx_error} characterizes the approximation capability of $\mathcal{F}(L,\mathbf{p},s,F)$. At the minimax sparsity level ($\delta=0$), we have $\eta=0$
so that 
\(
\|g_{\textrm{apx}}-g_0\|_\infty = O(\phi_n^{1/2}).
\)
For $\delta>0$, overparameterization sharpens this
 bound to $O(\phi_n^{(1+\eta)/2})$. We next establish a non-asymptotic
population risk bound for any estimator $\hat g \in \mathcal{F}(L,\mathbf{p},s,F)$.

\begin{theorem}\label{th:Global_risk}
Under Conditions \ref{as:nondegcox} and \ref{as:DNN_structures} for $0\leq \delta < c_{\text{eff}}$, for any estimator $\hat{g} \in \mathcal{F}(L,\mathbf{p},s,F)$, there exist constants $c^*, C_1^*, C_2^* > 0$, depending only on the network bound $F$ and the structural constants defined in Condition \ref{as:DNN_structures}, such that the population risk satisfies
\begin{align}
\max\Big\{0,\underbrace{\frac{1}{2}\mathcal{Q}_{n}(\hat{g},\widehat{g}_{\text{opt}})}_{\text{optimization gap}}& -\underbrace{c^*\phi_{n}n^\delta L\log^2(n)}_{\text{statistical error}}\Big\}  \nonumber \\ \leq &\mathbb{E}_{\mathcal{D}_n}[R(\hat{g},g_0)] \leq \underbrace{2\mathcal{Q}_{n}(\hat{g},\widehat{g}_{\text{opt}})}_{\text{optimization gap}} + \underbrace{C_1^*\|g_{\text{apx}} - g_0\|^2_\infty }_{\text{approximation error}}
+ \underbrace{C_2^*\phi_{n}n^\delta L\log^2(n)}_{\text{statistical error}}.\label{eq: Global risk}
\end{align}
\end{theorem}
   Theorem~\ref{th:Global_risk} establishes that the population risk is bounded by 
the optimization gap together with approximation and statistical error terms, 
and clarifies the role of $\delta$. When $\delta=0$ and the expected optimization 
error $\mathcal{Q}_n$ is small, the bound attains the minimax rate 
$O(\phi_n L \log^2 n)$ \citep{Zhong2022-um}. For $0<\delta<c_{\mathrm{eff}}$, 
the rate increases to $O(\phi_n n^\delta L \log^2 n)$ but still vanishes since 
$\phi_n n^\delta L \log^2 n = o(1)$.
While the oracle inequality provides global control of the population risk, its 
local Kullback--Leibler structure induces quadratic curvature 
(Lemma~3 in appendix), yielding pointwise error bounds in the following 
corollary and enabling valid inference after bias calibration.
\begin{cy}[Pointwise Mean Squared Error Bound]
\label{cor:pointwise_mse}
 Suppose the conditions of Theorem~\ref{th:Global_risk} hold with 
$0 \le \delta < c_{\text{eff}}$, and let $\hat g \in \mathcal{F}(L,\mathbf{p},s,F)$ 
satisfy
\(
\mathcal{Q}_n(\hat g,\widehat g_{\mathrm{opt}})
\le C_1^* \phi_n n^\delta L \log^2 n
\)
for some constant $C_1^*>0$. For any slowly diverging sequence 
$M_n \to \infty$, if $\mathbf{x}_* \sim \Prob_{\mathbf X}$ and is independent of $\mathcal D_n$, then the pointwise mean squared error satisfies
\[
\mathbb{E}_{\mathcal D_n}\Big[(\hat g(\mathbf{x}_*)-g_0(\mathbf{x}_*))^2\Big]
\le M_n \phi_n n^\delta L \log^2 n
\]
with probability at least $1-M_n^{-1}$ with respect to $\Prob_{\mathbf X}$.
\end{cy}

Let
\begin{align}
\mathcal{B}_{n,\delta}
= \Big\{\mathbf{x}_* \in [0,1]^{p_0} :
\mathbb{E}_{\mathcal{D}_n}\!\left[(\hat g(\mathbf{x}_*)-g_0(\mathbf{x}_*))^2\right]
\le M_n \phi_n n^\delta L \log^2 n
\Big\}
\label{eq:B_n,delta}
\end{align}
denote the subset characterized by Corollary~\ref{cor:pointwise_mse}, which satisfies 
$\Prob_{\mathbf X}(\mathcal{B}_{n,\delta}) \ge 1-M_n^{-1}$. 
For $\mathbf{x}_* \in \mathcal{B}_{n,\delta}$, the pointwise MSE vanishes when 
$0 \le \delta < c_{\text{eff}}$. 
To establish asymptotic normality at  $\mathbf{x}_*$, however, the 
pointwise bias must also decay at a suitable rate. 
Because the estimator $\widehat g^b$ in \eqref{eq: ESM g} aggregates base learners 
trained on subsamples via gradient-based optimization, both the statistical 
estimation bias and the algorithmic optimization bias need to be controlled at the 
subsample level.
  
%Under Condition~\ref{as:nondegcox} and by Lemma 2 of \cite{Zhong2022-um}, 
%\[
%\mathbb{E}_{\mathbf{X}, \mathcal{D}_n}\Big[\big(\hat{g}(\mathbf{X}) - g_0(\mathbf{X})\big)^2\Big] \asymp \mathbb{E}_{\mathcal{D}_n}\big[L_0(g_0)-L_0(\hat{g})\big] = R(\hat{g}, g_0) \lesssim \phi_n n^{\delta}L\log^2 n.
%\]
%By Fubini's theorem, this population risk is equivalent to the expectation of the pointwise mean squared error over the covariate distribution $P_{\mathbf{X}}$
%\begin{align}
%\mathbb{E}_{\mathbf{X}}\bigg[ \mathbb{E}_{\mathcal{D}_n}\big(\hat{g}(\mathbf{X}) - g_0(\mathbf{X})\big)^2 \bigg] \lesssim \phi_n n^\delta L \log^2 n.\label{eq:B_n,delta}
%\end{align}
%An application of Markov's inequality on $([0,1]^{p_0}, P_{\mathbf{X}})$ guarantees that for any slowly diverging sequence $M_n \to \infty$, the set 
%\[
%\mathcal{B}_{n, \delta} = \bigg\{ \mathbf{x}_* \in [0,1]^{p_0} : \mathbb{E}_{\mathcal{D}_n}\Big[\big(\hat{g}(\mathbf{x}_*) - g_0(\mathbf{x}_*)\big)^2\Big] \le M_n \phi_n n^\delta L \log^2 n \bigg\}
%\]

\begin{as}\label{as_bias_domination}

For each subsample $\mathcal{D}_r$ (indexed by $\mathcal{I}^b$) with size $r$, we assume the pointwise estimation bias of the trained estimator $\widehat{g}^b \in \mathcal{F}(L,\mathbf{p},s,F)$ defined in \eqref{eq:g_hat_SGD} is governed by the optimal  network $g_{\textrm{apx}}$. For any target point $\mathbf{x}_* \in [0,1]^{p_0}$,
\[
\big| \mathbb{E}_{\mathcal{D}_r}[\widehat{g}^b(\mathbf{x}_*)] - g_{\textrm{apx}}(\mathbf{x}_*) \big| \lesssim \{A_r(g_{\textrm{apx}}, \mathbf{x}_*)\}^{1-\xi_r},
\]
for a small $ 0< \xi_r < 1$, where $A_r(g_{\textrm{apx}}, \mathbf{x}_*) = |g_{\textrm{apx}}(\mathbf{x}_*) - g_0(\mathbf{x}_*)|$ is the pointwise approximation error of $g_{\textrm{apx}}$ at the subsample level $r$.
\end{as}

\begin{as}\label{as_opt_error}
For each subsample $\mathcal{D}_r$  (indexed by $\mathcal{I}^b$) with size $r$, the maximum optimization error $\mathcal{Q}^{\text{opt}}_r$ defined in \eqref{eq:SGC-opt} satisfies
\[
\mathcal{Q}^{\text{opt}}_r \lesssim \phi_r r^{\delta}L\log^2r.
\]
\end{as}

Condition~3 is anchored at the approximation benchmark 
$g_{\mathrm{apx}}$, the best achievable element in 
$\mathcal{F}(L,\mathbf{p},s,F)$. It requires that, on average, the 
trained estimator is close to $g_{\mathrm{apx}}$, thereby separating 
approximation error from training error. This formulation is intrinsic to the 
function class and does not impose an artificial centering at $g_0$. Any 
additional deviation reflects optimization deficiency rather than the 
approximation limit and cannot be absorbed into the bias term, so the condition 
is necessary to control bias at the scale required for asymptotic normality. As in classical nonparametric inference \citep{Wasserman2006}, asymptotic 
normality requires undersmoothing so that the squared bias is negligible 
relative to the variance. Under the minimax regime ($\delta=0$), 
$\|g_{\mathrm{apx}}-g_0\|_\infty = O(\phi_r^{1/2})$, yielding pointwise bias 
$O(\phi_r^{(1-\xi_r)/2})$, which is not negligible relative to the variance. 
Thus, an undersmoothed regime ($\delta>0$) is required, reflecting the same 
bias–variance tradeoff as in classical settings. Condition~\ref{as_opt_error} controls the optimization gap by requiring that 
the trained estimator attains the statistical scale of the oracle bound. This 
condition is stated in terms of the empirical objective actually optimized in 
practice and does not assume exact minimization. 
Without such control, the 
optimization error would enter the leading term of the risk and prevent valid 
bias calibration, making it  necessary for inference.

\begin{as}\label{as_xi}
Assume that $B \gtrsim n$, $r=n^{\alpha}$ for $0<\alpha<1$, and the single-overlap covariance satisfies 
\(
\inf_{\mathbf{x}_*\in [0,1]^{p_0}}\zeta^{(1)}_r(\mathbf{x}_*) \gtrsim r^{-\nu}
\) for some $\nu\geq 1$.
\end{as}
Condition~\ref{as_xi} imposes standard U-statistic scaling so that the 
first-order H\'ajek projection dominates the asymptotic expansion, and 
Monte Carlo variability from ensembling is asymptotically negligible. 
Because the DNN class admits a bounded envelope, the H\'ajek decomposition 
implies that the leading covariance is $O(1/r)$, yielding the natural lower bound 
$\nu \ge 1$. 
% Allowing $\nu$ to vary provides a weaker and more flexible 
% condition than that in \citet{Meng2025-zh}.

\begin{theorem}
\label{thm:asn-bagged}
  Suppose $g_0$ belongs to the class \eqref{eq:true_functional_class} with 
$g_0(\mathbf{0})=0$. Let $\hat g_B$ be the ensemble estimator defined in 
\eqref{eq: ESM g} based on $B$ subsamples of size $r$, and let 
$\mathcal{B}_{r,\delta}$ be defined in \eqref{eq:B_n,delta} with subsample size $r$. 
Assume Condition~\ref{as:DNN_structures} holds with hyper parameter
$\delta \in \big(\frac{c_{\mathrm{eff}}(1-c_{\mathrm{eff}})}{2-c_{\mathrm{eff}}}, 
c_{\mathrm{eff}}\big)$, and Conditions~\ref{as_bias_domination}--\ref{as_xi} hold 
at subsample size $r$. Let $m_0 = c_{\mathrm{eff}}(1+\eta)(1-\xi_r)$.  
Then, with $\nu \in \big(1, 1 + \frac{m_0}{2}\big)$,  for any $\alpha \in (\alpha_{\mathrm{lower}}, \alpha_{\mathrm{upper}})$ and any fixed 
$\mathbf{x}_* \in \mathcal{B}_{r,\delta}$, it holds that
\begin{equation*}
\sqrt{\frac{n}{r^2\zeta^{(1)}_{r}(\mathbf{x}_*)}} 
\big(\hat g_B(\mathbf{x}_*) - g_0(\mathbf{x}_*)\big) 
\xrightarrow{d} \mathcal{N}(0,1),
\end{equation*}
where
\[
\alpha_{\mathrm{lower}}=\frac{1}{2-\nu+m_0}, 
\qquad 
\alpha_{\mathrm{upper}}=\frac{1}{\nu+c_{\mathrm{eff}}-\delta}.
\]
\end{theorem}

This range reflects two competing requirements in ESM. The lower bound  $\alpha > \alpha_{\mathrm{lower}}$ arises from bias control: the subsample size  must be sufficiently large so that the  bias is asymptotically negligible. 
In contrast, the upper bound $\alpha < \alpha_{\mathrm{upper}}$ is imposed by the 
need for H\'ajek domination, requiring the variance of the remainder to vanish.
A similar tradeoff appears in random forests, where the single-overlap covariance 
decays at rate $1/\{r(\log r)^{p_0}\}$ (i.e., $\nu \approx 1$) \citep{WagerAthey2018}. 
In our setting, a nonempty admissible interval 
$(\alpha_{\mathrm{lower}},\alpha_{\mathrm{upper}})$ is ensured when $\nu < 1 + \frac{m_0}{2}$, which guarantees separation between the bias 
and variance constraints. The parameter $\delta$ governs this tradeoff through its effect on approximation: 
when $\delta$ is small, the approximation bias decays more slowly, increasing 
$\alpha_{\mathrm{lower}}$ and requiring larger subsamples; as $\delta$ increases, 
bias decays faster due to undersmoothing, but this enlarges the projection remainder 
variance, which decreases $\alpha_{\mathrm{upper}}$. Consequently, more aggressive 
undersmoothing narrows the admissible range for $\alpha$.

  At fixed evaluation points 
$\mathbf{x}_\ast^{(1)},\dots,\mathbf{x}_\ast^{(p)} \in [0,1]^{p_0}$, 
define the vectors of ensemble estimators and true function values as
\begin{equation} \label{gb}
\hat{\mathbf{g}}_{B} := 
\begin{pmatrix}
\hat g_{B}(\mathbf{x}_\ast^{(1)}) \\[-0.2em]
\vdots\\[-0.2em]
\hat g_{B}(\mathbf{x}_\ast^{(p)})
\end{pmatrix},
\qquad
\mathbf{g}_{0} := 
\begin{pmatrix}
g_0(\mathbf{x}_\ast^{(1)})\\[-0.2em]
\vdots\\[-0.2em]
g_0(\mathbf{x}_\ast^{(p)})
\end{pmatrix}.
\end{equation}

\begin{as}\label{as_multivar_cov}
    Let $\mathbf{\Psi}_r$ denote the $p\times p$ single-overlap covariance matrix evaluated at 
$\mathbf{x}_\ast^{(1)},\dots,\mathbf{x}_\ast^{(p)}$, with entries 
$(\mathbf{\Psi}_r)_{ij}=\zeta^{(1)}_{r}\!\big(\mathbf{x}_\ast^{(i)},\mathbf{x}_\ast^{(j)}\big)$ 
for $1\le i,j\le p$. With 
$\zeta^{(1)}_{r}(\mathbf{x}_*) = \zeta^{(1)}_{r}(\mathbf{x}_*,\mathbf{x}_*)$, 
define
\(
\mathbf{D}_r=\mathrm{diag}\Big\{\zeta^{(1)}_{r}\!\big(\mathbf{x}_\ast^{(1)}\big),\dots,
\zeta^{(1)}_{r}\!\big(\mathbf{x}_\ast^{(p)}\big)\Big\}.
\)
Let $\mathbf{R}_r=\mathbf{D}_r^{-1/2}\mathbf{\Psi}_r \mathbf{D}_r^{-1/2}$ denote the normalized 
single-overlap covariance matrix. We assume $\mathbf{R}_r$ is uniformly well-conditioned, i.e.,
$\lambda_{\min}(\mathbf{R}_r)\ge c_0>0$ for all sufficiently large $r$, where 
$\lambda_{\min}(\mathbf{R}_r)$ denotes the minimum eigenvalue of $\mathbf{R}_r$.
\end{as}
  Condition~\ref{as_multivar_cov} is a standard nondegeneracy condition ensuring 
that the estimates at the $p$ evaluation points are not perfectly correlated, 
thereby yielding a well-defined limiting multivariate normal distribution 
\citep{jacot2018neural}. We extend Theorem~\ref{thm:asn-bagged} to the multivariate 
setting by establishing asymptotic normality of $\hat{\mathbf g}_B$ in \eqref{gb}.
 \begin{theorem}
\label{thm:asn-bagged-multivar}
 Suppose that $g_0$ belongs to the class \eqref{eq:true_functional_class} with $g_0(\mathbf{0}) = 0$. Let $\hat{\mathbf{g}}_B$ and $\mathbf{g}_0$ be defined as in \eqref{gb} based on $B$ subsamples of size $r$, $\mathcal{B}_{r,\delta}$ be as defined in \eqref{eq:B_n,delta} with sample size $r$ and hyper parameter $\delta \in \Big(\frac{c_{\mathrm{eff}}(1-c_{\mathrm{eff}})}{2-c_{\mathrm{eff}}}, c_{\mathrm{eff}}\Big)$. Assume Conditions~\ref{as_bias_domination}--\ref{as_multivar_cov} hold. Then, for any $\nu \in \big(1, 1 + \frac{m_0}{2}\big)$, there exist $\alpha_{\mathrm{lower}}$ and $\alpha_{\mathrm{upper}}$ such that for all $\alpha \in (\alpha_{\mathrm{lower}}, \alpha_{\mathrm{upper}})$ and any fixed points $\mathbf{x}_*^{(1)},\dots,\mathbf{x}_*^{(p)} \in \mathcal{B}_{r,\delta}$,
\begin{equation}
\label{eq:ASN_bag_multivar_matrix}
\sqrt{\frac{n}{r^2}}\,
\mathbf{\Psi}_r^{-1/2}\,(\hat{\mathbf{g}}_{B}-\mathbf{g}_{0})
\xrightarrow{d} \mathcal{N}_p(\mathbf{0}, \mathbf{I}_p),
\end{equation}
where $\alpha_{\mathrm{lower}}$ and $\alpha_{\mathrm{upper}}$ are the same as defined in Theorem \ref{thm:asn-bagged}.
\end{theorem}
For pairwise contrasts, consider the special case $p=2$ evaluated at $\mathbf{x}_\ast^{(1)}$ and $\mathbf{x}_\ast^{(2)}$. 
Taking $v = (1,-1)^\top$, the corresponding pairwise variance is
\begin{align*}
\sigma_r^2\bigl(\mathbf{x}_\ast^{(1)},\mathbf{x}_\ast^{(2)}\bigr)
&:= v^\top \mathbf{\Psi}_r v \\
&= \zeta^{(1)}_r(\mathbf{x}_\ast^{(1)}) + \zeta^{(1)}_r(\mathbf{x}_\ast^{(2)})
- 2\,\zeta^{(1)}_r\bigl(\mathbf{x}_\ast^{(1)},\mathbf{x}_\ast^{(2)}\bigr).
\end{align*}
Because Condition~\ref{as_multivar_cov} for $p=2$ ensures the $2\times 2$ covariance matrix is uniformly well-conditioned, the contrast variance is non-degenerate. 
We define the corresponding scalar normalizing factor as
\[
b_{n,r}\bigl(\mathbf{x}_\ast^{(1)},\mathbf{x}_\ast^{(2)}\bigr)
:=
\sqrt{\frac{n}{r^{2}\,\sigma_r^2\bigl(\mathbf{x}_\ast^{(1)},\mathbf{x}_\ast^{(2)}\bigr)}}.
\]

\begin{cy}
\label{cor:asn-bagged-contrast}
Under the conditions of Theorem~\ref{thm:asn-bagged-multivar} with $p=2$, for any fixed pair of covariate points $\mathbf{x}_\ast^{(1)}, \mathbf{x}_\ast^{(2)}$ and any subsample scaling exponent satisfying $\alpha_{\mathrm{lower}} < \alpha < \alpha_{\mathrm{upper}}$,
\begin{equation*}
b_{n,r}\bigl(\mathbf{x}_\ast^{(1)},\mathbf{x}_\ast^{(2)}\bigr)\,
\Bigl\{
\widehat{\psi}_{B}(\mathbf{x}_\ast^{(1)},\mathbf{x}_\ast^{(2)})
-
\psi(\mathbf{x}_\ast^{(1)},\mathbf{x}_\ast^{(2)})
\Bigr\}
\xrightarrow{d} 
\mathcal{N}(0,1).
\end{equation*}
\end{cy}

Since analytical forms for $\zeta_r^{(1)}(\mathbf{x}_\ast)$ and 
$\zeta_r^{(1)}(\mathbf{x}_\ast^{(1)},\mathbf{x}_\ast^{(2)})$ are unavailable, 
the next theorem provides their consistent estimators. 
\begin{theorem}
\label{thm:var-consistency}
Let $\mathcal{B}_{r,\delta}$  be  as defined in \eqref{eq:B_n,delta} with sample size $r$. For any two fixed evaluation points $\mathbf{x}^{(1)}_\ast, \mathbf{x}^{(2)}_\ast \in \mathcal{B}_{r,\delta}$, the following results hold.

(i) Under Condition~\ref{as_xi}, let $\hat{\sigma}^2(\mathbf{x}^{(k)}_*)$ denote the IJ variance estimator for the individual ensemble estimator $\widehat g_{B}(\mathbf{x}^{(k)}_*)$, as defined in \eqref{eq:sigma_pointwise}. Then, for any subsample scaling exponent satisfying {$1/2 < \alpha < \alpha_{\mathrm{upper}}$}, and for $k=1,2$ it holds
\begin{align}
\frac{n\,\hat{\sigma}^2(\mathbf{x}^{(k)}_*)}{r^{2}\,\zeta_r^{(1)}(\mathbf{x}^{(k)}_{*})} \xrightarrow{p} 1.
\label{eq:i-consistency}
\end{align}

(ii) Suppose, in addition, that Condition~\ref{as_multivar_cov} holds. Let $\hat{\sigma}^2_{1,2}$ denote the IJ variance estimator for the contrast $\widehat g_{B}(\mathbf{x}^{(1)}_*) - \widehat g_{B}(\mathbf{x}^{(2)}_*)$, as defined in \eqref{variance est}. Then, for any subsample scaling exponent satisfying {$1/2 < \alpha < \alpha_{\mathrm{upper}}$},
\begin{align}
\frac{n\,\hat{\sigma}^2_{1,2}}{r^{2}\,\sigma_r^2\bigl(\mathbf{x}_\ast^{(1)},\mathbf{x}_\ast^{(2)}\bigr)} \xrightarrow{p} 1.
\label{eq:ij-consistency}
\end{align}
\end{theorem}
Theorem~\ref{thm:var-consistency} shows that when $B \gtrsim n$ and $1/2 < \alpha < \alpha_{\mathrm{upper}}$, the IJ estimator consistently estimates the covariance of the first-order H\'ajek projection after correcting for Monte Carlo variability from subsampling.

%\begin{remark}[Roles of Conditions 3--6]
%Conditions~3--5 characterize the key requirements for pointwise inference. 
%Condition~3 controls the pointwise bias, requiring it to be asymptotically 
%negligible relative to the stochastic fluctuation, analogous to undersmoothing 
%in classical nonparametric inference. Condition~4 controls the optimization gap 
%from gradient-based training, requiring that the estimator attains the statistical 
%scale of the oracle bound without exact empirical risk minimization. Condition~5 
%governs the dependence induced by subsampling and ensures dominance of the 
%first-order H\'ajek projection, which underlies asymptotic normality and IJ consistency. Condition~6 ensures that the limiting covariance matrix is nondegenerate, which  is required for multivariate normality and contrast inference.  These conditions are not needed for global risk consistency, but are required for distributional approximations of pointwise and contrast functionals.
%\end{remark}

%For relative risk inference, this is essential because the errors in $g_0(\mathbf{x}_*^{(1)})$ and $g_0(\mathbf{x}_*^{(2)})$ propagate  to the log-hazard ratio. Theorem~\ref{thm:var-consistency} therefore guarantees valid inference for $\hat g_{B}$ and the corresponding log-hazard ratio over all admissible subsample sizes $r=n^\alpha$.
 
%%%%%%%%%%%%%%%%%%%%%%%%%%%%%%%%%%%%%%%%%%%%%%%%%%%%%%%%%%%%%%%%%%%%%%%%%%%%
%% SIMULATION STUDIES (COMPRESSED DROP-IN VERSION)
%%%%%%%%%%%%%%%%%%%%%%%%%%%%%%%%%%%%%%%%%%%%%%%%%%%%%%%%%%%%%%%%%%%%%%%%%%%%

\section{Simulation Studies}
\label{sec:4}
% \ylcm{expand simulations by empirically validating subsample window;  a bias-dominance failure experiment; IJ variance scaling and B threshold; Scaling in higher dimension, eg $p$ is large}
 % \sgcm{I am running, but problem is we have to do 200 jobs parallely for varying $r=0.65,0.70,..0.95,0.98$ 8 indices for $r$ and varying drop out $0.2$ to $0.5$ which will take $200*2*8=3200$ jobs altogether, so might take some time. For now I'm running for only $0.65,0.75,0.85,0.95$ with drop out $0.5$ to see. }  \sgcm{Sorry Dr Li, unfortunately still I need to wait until tomorrow to get final approval from AR since its past 5pm to continue the remaining simulations.}
    Simulation studies assess the finite-sample performance of the proposed ESM estimator for the nonparametric risk function \(g_0(\mathbf{x}_*)\) and the log-hazard ratio \(g_0(\mathbf{x}_*^{(1)}) - g_0(\mathbf{x}_*^{(2)})\). For each training dataset of size \(n\), covariates \(\mathbf{X}_i\) are generated i.i.d.\ from \(N(\mathbf{0}, I_{10})\) and truncated componentwise to \([-3,3]\), for \(i=1,\ldots,n\). The function \(g_0(\cdot)\) depends on the first three components of \(\mathbf{X}_i\), with the remaining seven serving as noise variables. Conditional on \(\mathbf{X}_i\), event times \(T_{i,U}\) are generated from the Cox model:
\[
\lambda(t\mid\mathbf{X}_i)=0.1\,t\exp\{g_0(\mathbf{X}_i)\},
\]
while censoring times \(T_{i,C}\) are generated independently from an exponential distribution with the rate chosen to yield approximately \(30\%\) censoring. For various levels of functional complexity, we consider three forms of \(g_0\).
\medskip

\noindent\textbf{Case 1 (Linear).}
\[
g_{0,1}(\mathbf{\mathbf{x}})= x_1 - 1.2\,x_2 + 0.8\,x_3.
\]

\medskip
\noindent\textbf{Case 2 (Smooth additive nonlinear).}
\[
 g_{0,2}(\mathbf{x})= 0.7 x_1 - 0.5\,x_3 + 0.4\,x_2^2
        + 0.3\,x_1 x_2.
\]

\medskip
\noindent\textbf{Case 3 (Compositional nonlinear).}
\begin{align*}
g_{0,3}(\mathbf{x})
=&\,0.7 x_1
- 0.8 x_3
+ 0.5 x_2^2
+ \sin(0.5 x_1 x_2) \\
&\quad + 1.2 \exp\!\Big(-0.25(x_1-1)^2 - 0.25(x_2+1)^2\Big)
- 1.2 \exp(-0.5).
\end{align*}

\begin{figure}[htp]
  \centering
  \begin{subfigure}[t]{0.48\textwidth}
    \centering
    \includegraphics[width=\textwidth]{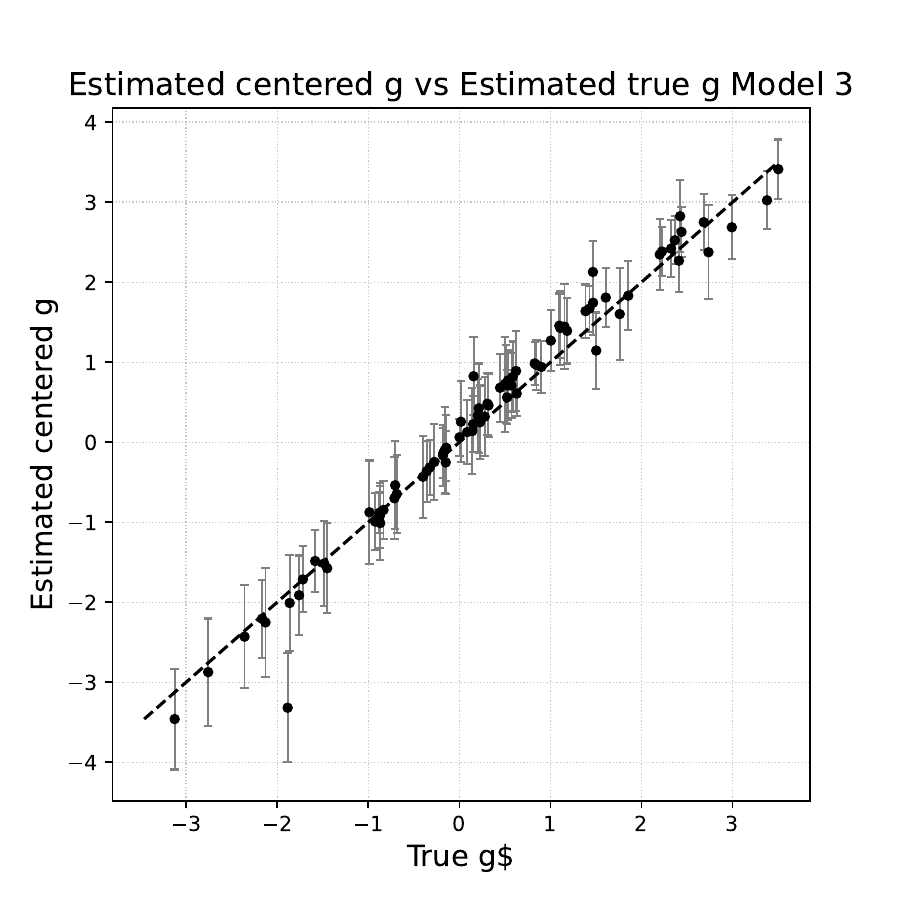}
    \caption{Point estimates and variations.}
    \label{fig:m3_point_est}
  \end{subfigure}
  \hfill
  \begin{subfigure}[t]{0.46\textwidth}
    \centering
    \includegraphics[width=\textwidth]{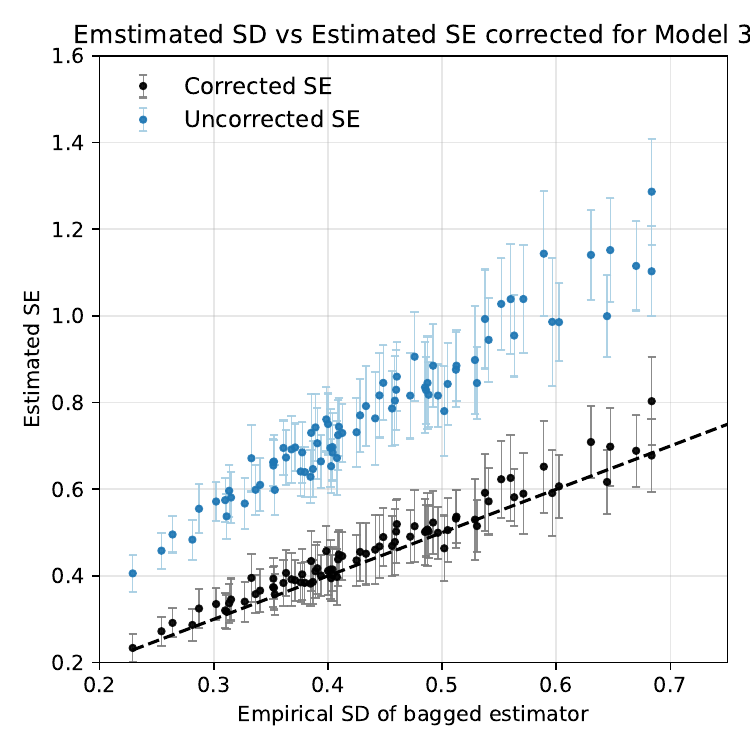}
    \caption{Estimated standard errors (corrected vs.\ uncorrected).}
    \label{fig:m3_se_diag}
  \end{subfigure}

  \caption{Estimation and inference summary for Model~3 with $n=800$, $r=\lfloor n^{0.90}\rfloor$, $B=1000$,
  DNN architecture $(p_0,128,64,1)$. Panel (a) plots $\hat g_{B}(\mathbf{x})$ versus $g_0(\mathbf{x})$ across test points.
  Panel (b) compares $\EmpSD$ with the estimated $\SE$, highlighting corrected and uncorrected versions when available.}
  \label{fig:sim_model3_example}
\end{figure}

  An independent test set $\{\mathbf{x}_{\mathrm{test},i}\}_{i=1}^{80}$ is generated i.i.d.\ from $N(\mathbf{0}, I_{10})$, truncated componentwise to $[-3,3]$, and held fixed across replications. For each training dataset, $\hat g_{B}(\mathbf{x}_{\mathrm{test},i})$ and its bias-corrected IJ variance estimate are computed. The settings are
\[
n \in \{800,\,1000\},\qquad r=\lfloor n^\alpha \rfloor,\quad \alpha\in\{0.85,\,0.90,\,0.95\},
\]
with 200 Monte Carlo replications per configuration. The log partial likelihood in~\eqref{eq: g_hat DNN} is optimized in PyTorch \citep{10.5555/3454287.3455008}. Each ensemble uses $B=1000$ subsamples, and each base learner is a DNN with architecture $(p_0,128,64,1)$, weight decay 0.02, learning rate 0.1, 500 epochs, and dropout 0.1, with $p_0=10$.
\begin{table}[htp]
\centering
\begin{threeparttable}
\caption{Simulation summary for pointwise inference on $g_0(\mathbf{x})$ across test points.  EmpSD is the empirical standard deviation of $\hat g_{B}(\mathbf{x})$ over replications.
SE is the mean estimated standard error.
CP is the empirical coverage probability of nominal 95\% intervals, and AIL is the mean interval length.}

\label{tab:sim_g_pointwise}

\small
\setlength{\tabcolsep}{3pt}
\renewcommand{\arraystretch}{1.1}

\begin{tabular}{c c c c c c c c}
\toprule
Case & Setting & Bias & MAE & EmpSD & $\textrm{SE}$ & CP & $\textrm{AIL}$ \\
\midrule

\multirow{3}{*}{1}
 & $n=800,\ r=\lfloor n^{0.85}\rfloor$ & -0.02 & 0.36 & 0.48 & 0.50 & 0.87 & 1.98 \\
 & $n=800,\ r=\lfloor n^{0.90}\rfloor$ & -0.02 & 0.21 & 0.45 & 0.48 & 0.92 & 1.86 \\
 & $n=800,\ r=\lfloor n^{0.95}\rfloor$ & -0.03 & 0.11 & 0.40 & 0.44 & 0.95 & 1.72 \\
\addlinespace
\midrule
\multirow{3}{*}{2}
 & $n=800,\ r=\lfloor n^{0.85}\rfloor$ & 0.11 & 0.26 & 0.47 & 0.49 & 0.89 & 1.91 \\
 & $n=800,\ r=\lfloor n^{0.90}\rfloor$ & 0.05 & 0.14 & 0.42 & 0.44 & 0.93 & 1.74 \\
 & $n=800,\ r=\lfloor n^{0.95}\rfloor$ & 0.04 & 0.11 & 0.36 & 0.40 & 0.95 & 1.59  \\
\addlinespace
\midrule
\multirow{3}{*}{3}
 %& $n=800,\ r=\lfloor n^{0.80}\rfloor$ &0.02 &0.42 &0.51 &0.52 &0.84 &2.05 \\
 & $n=800,\ r=\lfloor n^{0.85}\rfloor$ & 0.01 & 0.28 & 0.48 & 0.49 & 0.89 & 1.93 \\
 & $n=800,\ r=\lfloor n^{0.90}\rfloor$ & 0.05 & 0.17 & 0.44 & 0.46 & 0.92 & 1.81  \\
 & $n=800,\ r=\lfloor n^{0.95}\rfloor$ & 0.02 & 0.14 & 0.40 & 0.43 & 0.94 & 1.68 \\
\midrule

\multirow{3}{*}{1}
 %& $n=1000,\ r=\lfloor n^{0.80}\rfloor$ &0.00 &0.25 &0.40 &0.40 &0.87 &1.57 \\
 & $n=1000,\ r=\lfloor n^{0.85}\rfloor$ & -0.03 & 0.26 & 0.41 & 0.42 & 0.89 & 1.66 \\
 & $n=1000,\ r=\lfloor n^{0.90}\rfloor$ & -0.05 & 0.15 & 0.37 & 0.39 & 0.93 & 1.52 \\
 & $n=1000,\ r=\lfloor n^{0.95}\rfloor$ & -0.03 & 0.06 & 0.34 & 0.36 & 0.95 & 1.40 \\
\addlinespace
\midrule
\multirow{3}{*}{2}
 & $n=1000,\ r=\lfloor n^{0.85}\rfloor$ & 0.09 & 0.19 & 0.39 & 0.40 & 0.91 & 1.58\\
 & $n=1000,\ r=\lfloor n^{0.90}\rfloor$ & 0.04 & 0.10 & 0.35 & 0.36 & 0.94 & 1.41 \\
 & $n=1000,\ r=\lfloor n^{0.95}\rfloor$ & 0.06 & 0.12 & 0.31 & 0.33 & 0.93 & 1.29 \\
\addlinespace
\midrule
\multirow{3}{*}{3}
 & $n=1000,\ r=\lfloor n^{0.85}\rfloor$ & 0.02 & 0.23 & 0.41 & 0.42 & 0.89 & 1.64 \\
 & $n=1000,\ r=\lfloor n^{0.90}\rfloor$ & -0.00 & 0.12 & 0.37 & 0.38 & 0.92 & 1.49 \\
 & $n=1000,\ r=\lfloor n^{0.95}\rfloor$ & 0.03 & 0.15 & 0.33 & 0.35 & 0.92 & 1.40 \\
\bottomrule
\end{tabular}
% \begin{tablenotes}[flushleft]
% \footnotesize
% \item EmpSD is the empirical standard deviation of $\hat g_{\textrm{bag}}(\mathbf{x})$ over replications.
% SE is the mean estimated standard error.
% CP is the empirical coverage probability of nominal 95\% intervals, and AIL is the mean interval length.
% \end{tablenotes}
\end{threeparttable}
\end{table}
  For pointwise inference on $g_0(\mathbf{x}_{\mathrm{test},i})$, performance is summarized by bias, mean absolute error (MAE), empirical standard deviation (EmpSD), corrected IJ standard error ($\SE$), coverage probability (CP), and corrected average interval length ($\AIL$), averaged over the test points.
For contrast inference, we consider
\(
\psi(\mathbf{x}_{\mathrm{test},i},\mathbf{x}_{\mathrm{test},j})
=
g_0(\mathbf{x}_{\mathrm{test},i})-g_0(\mathbf{x}_{\mathrm{test},j}),
\)
where $(i,j)$ index two distinct test points selected from the same test set.
We focus on pairs satisfying
\(
g_0(\mathbf{x}_{\mathrm{test},i}),\, g_0(\mathbf{x}_{\mathrm{test},j}) \in [-1.5, 2.0]
\quad\text{and}\, \,
|g_0(\mathbf{x}_{\mathrm{test},i}) - g_0(\mathbf{x}_{\mathrm{test},j})|\approx 1.0,
\)
which represent moderate signal levels and contrasts commonly encountered in the design.
We report the same metrics for $\hat{\psi}$. Figure~\ref{fig:sim_model3_example}
shows a representative visualization for Case~3.
Table~\ref{tab:sim_g_pointwise} summarizes pointwise inference, and
Table~\ref{tab:sim_loghr} summarizes contrast inference.

% -------------------- FIGURE (2 panels) --------------------
 
Overall, the pointwise bias is small relative to stochastic variation. 
For moderately large $\alpha$ (e.g., $\alpha \in [0.90, 0.95]$), 
the corrected IJ standard error closely matches the empirical standard deviation, 
yielding near-nominal coverage. In this regime, $\alpha$ is large enough to 
suppress approximation bias ($\alpha > \alpha_{\mathrm{lower}}$) while remaining 
below the threshold required for H\'ajek domination and asymptotic normality 
($\alpha < \alpha_{\mathrm{upper}}$).

\begin{figure}[htbp]
    \centering
    % Top Row: CP and MAE
    \begin{subfigure}{0.48\textwidth}
        \centering
        \includegraphics[width=\linewidth]{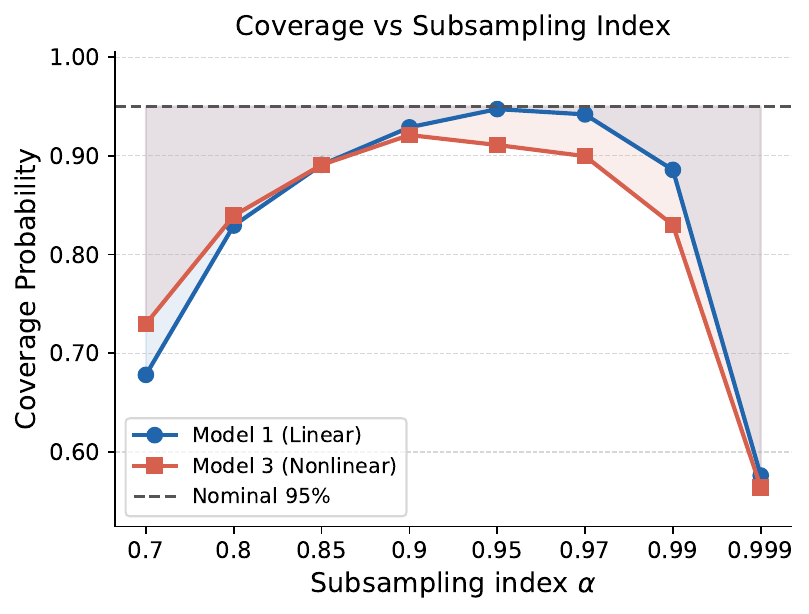}
        \caption{Coverage Probability}
    \end{subfigure}\hfill
    \begin{subfigure}{0.48\textwidth}
        \centering
        \includegraphics[width=\linewidth]{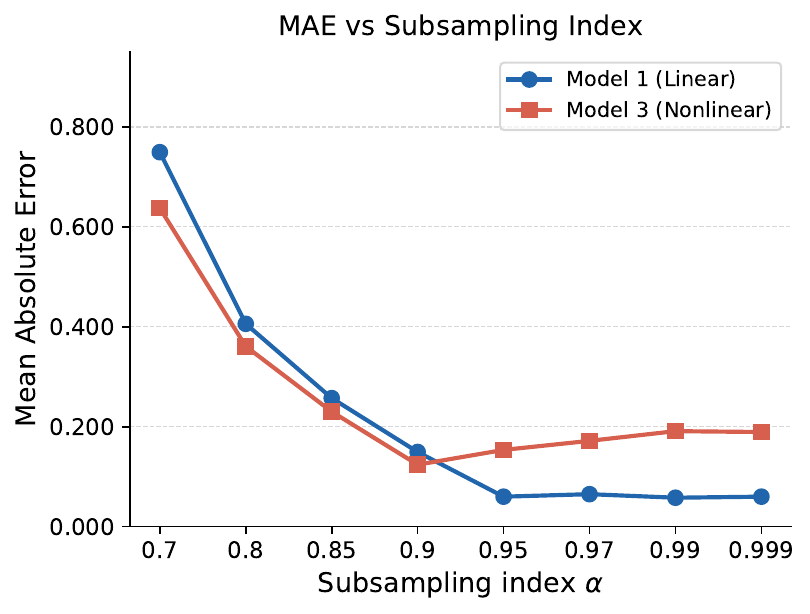}
        \caption{Mean Absolute Error}
    \end{subfigure}
    
    \vspace{0.3cm} % Minimal vertical spacing between rows
    
    % Bottom Row: SE vs EmpSD for both models
    \begin{subfigure}{0.48\textwidth}
        \centering
        \includegraphics[width=\linewidth]{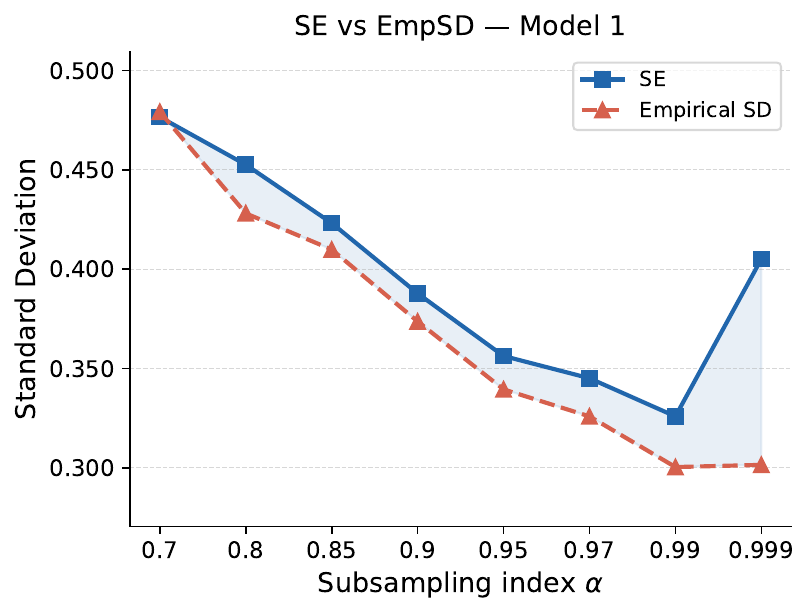}
        \caption{SE vs EmpSD (Model 1)}
    \end{subfigure}\hfill
    \begin{subfigure}{0.48\textwidth}
        \centering
        \includegraphics[width=\linewidth]{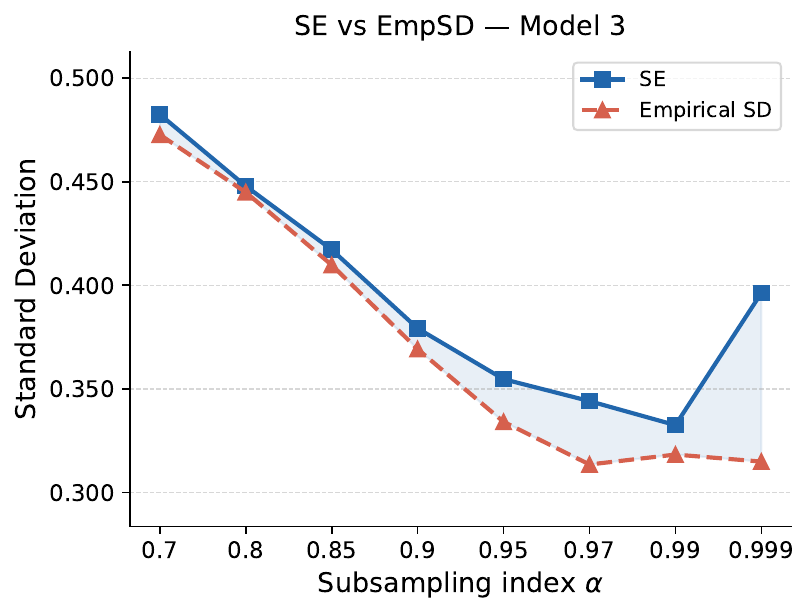}
        \caption{SE vs EmpSD (Model 3)}
    \end{subfigure}
    
    \caption{Empirical performance across the subsampling index $\alpha$. At smaller values ($\alpha \le 0.8$), high MAE degrades the CP, validating the theoretical lower bound $\alpha_{\text{lower}}$. Conversely, as $\alpha \to 1$, the CP drops significantly despite stable MAE, and variance estimates diverge, empirically validating the upper bound $\alpha_{\text{upper}}$.}
    \label{fig:alpha_upper_failure}
\end{figure}

 Figure~\ref{fig:alpha_upper_failure} illustrates the full empirical spectrum. 
For smaller values ($\alpha \le 0.8$), coverage deteriorates due to elevated 
approximation bias, consistent with the theoretical lower bound. In contrast, 
as $\alpha$ approaches 1, coverage collapses despite relatively stable MAE. 
This breakdown reflects the failure of the asymptotic Gaussian approximation 
as subsamples become highly overlapping. Correspondingly, the corrected standard 
error deviates from the empirical standard deviation in this regime, providing 
empirical evidence for the necessity of the upper bound 
$\alpha_{\mathrm{upper}}$.

Finally, increasing the sample size from $n=800$ to $n=1000$ improves 
estimation efficiency, as reflected by reductions in the empirical 
standard deviation (EmpSD) and the average interval length. In addition, 
contrast inference achieves coverage closer to the nominal level than 
pointwise inference. This improvement is attributable to partial bias 
cancellation in the contrast, and remains evident even under the strongly 
nonlinear Model~3.

% -------------------- TABLE: pointwise g --------------------

% -------------------- TABLE: log-HR contrast --------------------
\begin{table}[htp]
\centering
\begin{threeparttable}
\caption{Simulation summary for inference on the log-hazard ratio contrast
$\psi(\mathbf{x}_{\textrm{test},i},\mathbf{x}_{\textrm{test},j})$.}
\label{tab:sim_loghr}

\small
\setlength{\tabcolsep}{5pt}
\begin{tabular}{c c c c c c c c}
\toprule
Case & \hspace{12mm}Setting & Bias & MAE & EmpSD & SE & CP & AIL \\
\midrule

\multirow{3}{*}{1}
 & $n=800,\ r=\lfloor n^{0.85}\rfloor$ & -0.45 & 0.65 & 0.69 & 0.72 & 0.94 & 2.83\\
 & $n=800,\ r=\lfloor n^{0.90}\rfloor$ & -0.23 & 0.52 & 0.62 & 0.67 & 0.95 & 2.61 \\
 & $n=800,\ r=\lfloor n^{0.95}\rfloor$ & -0.22 & 0.50 & 0.55 & 0.63 & 0.97 & 2.46 \\
\addlinespace

\multirow{3}{*}{2}
 & $n=800,\ r=\lfloor n^{0.85}\rfloor$ & -0.17 & 0.59 & 0.71 & 0.69 & 0.94 & 2.72 \\
 & $n=800,\ r=\lfloor n^{0.90}\rfloor$ & -0.02 & 0.45 & 0.55 & 0.64 & 0.97 & 2.51\\
 & $n=800,\ r=\lfloor n^{0.95}\rfloor$ & 0.06 & 0.43 & 0.53 & 0.56 & 0.96 & 2.21 \\
\addlinespace

\multirow{3}{*}{3}
 & $n=800,\ r=\lfloor n^{0.85}\rfloor$ & -0.22 & 0.41 & 0.47 & 0.49 & 0.92 & 1.92 \\
 & $n=800,\ r=\lfloor n^{0.90}\rfloor$ & -0.21 & 0.36 & 0.41 & 0.47 & 0.93 & 1.85 \\
 & $n=800,\ r=\lfloor n^{0.95}\rfloor$ & -0.10 & 0.33 & 0.40 & 0.46 & 0.95 & 1.80 \\
\midrule

\multirow{3}{*}{1}
 & $n=1000,\ r=\lfloor n^{0.85}\rfloor$ & -0.22 & 0.53 & 0.58 & 0.60 & 0.92 & 2.35 \\
 & $n=1000,\ r=\lfloor n^{0.90}\rfloor$ & -0.28 & 0.51 & 0.56 & 0.55 & 0.91 & 2.16 \\
 & $n=1000,\ r=\lfloor n^{0.95}\rfloor$ & -0.18 & 0.41 & 0.48 & 0.51 & 0.96 & 1.97 \\
\addlinespace

\multirow{3}{*}{2}
 & $n=1000,\ r=\lfloor n^{0.85}\rfloor$ & -0.06 & 0.41 & 0.50 & 0.57 & 0.97 & 2.24 \\
 & $n=1000,\ r=\lfloor n^{0.90}\rfloor$ & 0.05 & 0.40 & 0.50 & 0.50 & 0.95 & 1.98 \\
 & $n=1000,\ r=\lfloor n^{0.95}\rfloor$ & 0.15 & 0.38 & 0.44 & 0.46 & 0.95 & 1.80  \\
\addlinespace

\multirow{3}{*}{3}
 & $n=1000,\ r=\lfloor n^{0.85}\rfloor$ & -0.20 & 0.37 & 0.41 & 0.41 & 0.92 & 1.61 \\
 & $n=1000,\ r=\lfloor n^{0.90}\rfloor$ & -0.04 & 0.31 & 0.38 & 0.40 & 0.94 & 1.55 \\
 & $n=1000,\ r=\lfloor n^{0.95}\rfloor$ & 0.02 & 0.28 & 0.35 & 0.38 & 0.96 & 1.48 \\
\bottomrule
\end{tabular}
% \begin{tablenotes}[flushleft]
% \footnotesize
% \item ${\psi}$ denotes the log-hazard ratio contrast $g_0(\mathbf{x}_{\textrm{test},i}) - g_0(\mathbf{x}_{\textrm{test},j})$.
% CP$_\psi$ and AIL$_\psi$ are computed for nominal 95\% Wald intervals for $\psi$.
% \end{tablenotes}
\end{threeparttable}
\end{table}

%%%%%%%%%%%%%%%%%%%%%%%%%%%%%%%%%%%%%%%%%%%%%%%%%%%%%%%%%%%%%%%%%%%%%%%%%%%%
%% BASELINE COMPARISON TABLE (AS PROVIDED)
%%%%%%%%%%%%%%%%%%%%%%%%%%%%%%%%%%%%%%%%%%%%%%%%%%%%%%%%%%%%%%%%%%%%%%%%%%%%

\subsection{Comparison with Competing Methods}
\label{subsec:sim_baselines}

\begin{table}[ht]
\centering
\caption{Estimation Performance and Uncertainty Quantification (Replications $R=200$, $\textrm{Bootstrap}= 200$) for Model 1 and 3. Pointwise metrics are averaged over the test set $x_{test}$.}
\label{tab:main_results}
\small
\begin{tabular}{cclccccc}
\toprule
Case & $n$ & Method & MAE & EmpSD & SE & CP & AIL \\ 
\midrule
\multirow{10}{*}{Case 1} & \multirow{5}{*}{800} & CoxPH & 0.11  &0.03 &0.14 &0.95  &0.57 \\
& & RSF &0.69  &0.02  &0.20 &0.53 &1.03\\
& & GBR &0.33  &0.03  &0.35 &0.89 &1.37\\
& & DeepSurv &0.39 &0.06 &2.52 &0.99 &2.63\\
& & \textbf{ESM} &0.11 &0.40 &0.44 &0.95 &1.72\\
 \cmidrule(lr){2-8} 
 & \multirow{5}{*}{1000} & CoxPH & 0.10 & 0.03 & 0.13 & 0.95 & 0.50 \\
 & & RSF       & 0.67 & 0.02 & 0.26 & 0.54 & 1.01 \\
 & & GBR      & 0.32 & 0.03 & 0.32 & 0.88 & 1.24 \\
 & & DeepSurv & 0.31 & 0.11 & 0.58 & 1.00 & 2.25 \\
 & & \textbf{ESM} & 0.06 & 0.34 & 0.36 & 0.95 & 1.40 \\
\midrule 
\multirow{10}{*}{Case 3} & \multirow{5}{*}{800} & CoxPH & 0.60 & 0.02 & 0.15 & 0.31 & 0.58 \\
 & & RSF      & 0.62 & 0.03 & 0.30 & 0.60 & 1.18 \\
 & & GBR      & 0.34 & 0.03 & 0.34 & 0.87 & 1.32 \\
 & & DeepSurv & 0.37 & 0.05 & 0.68 & 0.99 & 2.62 \\
 & & \textbf{ESM} & 0.14 & 0.40 & 0.43 & 0.94 & 1.68 \\ 
 \cmidrule(lr){2-8} 
 & \multirow{5}{*}{1000} & CoxPH & 0.60 & 0.01 & 0.13 & 0.26 & 0.52 \\
 & & RSF      & 0.60 & 0.02 & 0.30 & 0.60 & 1.16 \\
 & & GBR      & 0.32 & 0.03 & 0.30 & 0.86 & 1.20 \\
 & & DeepSurv & 0.30 & 0.04 & 0.57 & 0.99 & 2.24 \\
 & & \textbf{ESM} & 0.15 & 0.33 & 0.35 & 0.92 & 1.40 \\ 
\bottomrule
\end{tabular}
\end{table}

  ESM (with $r=\lfloor n^{0.95}\rfloor$) is compared with the Cox proportional hazards model (CoxPH), Random Survival Forest (RSF), Gradient Boosting Regression (GBR), and DeepSurv. Table~\ref{tab:main_results} summarizes pointwise MAE and the associated uncertainty quantification.
   Under the linear setting ({\bf Case~1}), CoxPH performs well, consistent with correct model specification. ESM remains competitive, achieving coverage close to the nominal 95\% level with relatively short intervals. 
Under the nonlinear setting ({\bf Case~3}), where the linear structure is misspecified, the advantage of ESM becomes pronounced. CoxPH maintains short intervals but exhibits reduced coverage, indicating unreliable uncertainty quantification. RSF and GBR attain lower MAE but tend to undercover, suggesting unstable variance estimation. DeepSurv achieves high coverage at the expense of wider intervals. 
In contrast, ESM attains lower MAE than competing methods while maintaining coverage near the nominal level with shorter intervals than DeepSurv, yielding a favorable accuracy–uncertainty tradeoff, particularly in the nonlinear setting.

% CoxPH performs well 
% under the linear model but exhibits poor coverage in nonlinear settings. 
% RSF and GBR achieve lower MAE but tend to undercover, indicating unstable 
% variance estimation. DeepSurv attains high coverage, but its intervals 
% are overly conservative. In contrast, ESM achieves coverage close to the 
% nominal 95\% level with shorter intervals, particularly under the nonlinear 
% Case~3, indicating a more favorable accuracy--uncertainty tradeoff. 
% An additional advantage is valid contrast inference: unlike competing  methods, ESM provides analytic covariance estimates between predictions  at different covariate values, enabling Wald-type confidence intervals for log-hazard ratio contrasts. 

%%%%%%%%%%%%%%%%%%%%%%%%%%%%%%%%%%%%%%%%%%%%%%%%%%%%%%%%%%%%%%%%%%%%%%%%%%%%
%% REAL DATA EXPERIMENTS (AS PROVIDED, LIGHTLY CLEANED)
%%%%%%%%%%%%%%%%%%%%%%%%%%%%%%%%%%%%%%%%%%%%%%%%%%%%%%%%%%%%%%%%%%%%%%%%%%%%

\section{Real Data Analysis}
\label{sec:realdata}  

  We analyze data from the Boston Lung Cancer Survival Cohort, restricted to patients with
non-small cell lung cancer \citep{wang2023blcsc}. The outcome is overall survival, defined as the time from diagnosis to death, with right-censoring at the last follow-up; the event indicator is recorded accordingly. Covariates include sex, age at diagnosis, smoking history (pack-years), tumor size (cm), chemotherapy status, cancer stage, and BMI. The cohort is randomly split into a 70\% training set ($n=1{,}454$) and a 30\% inference set ($n=623$).

The proposed ESM is compared with the Cox proportional hazards model (CoxPH),
Random Forest (RF), Gradient Boosting Regression (GBR), and a standalone DeepSurv network.
The ensemble uses $B=1000$ subsamples with subsample size $r=\lfloor n^{0.90}\rfloor$,
a dropout rate of 10\%, and a DNN architecture $(p_0,128,64,1)$. As reported in Table~\ref{tab:ESM-performance}, the ensemble achieves the highest mean
C-index and AUC, while maintaining the smallest standard errors among all methods.
It also yields the shortest average interval lengths (AIL) for both metrics, indicating
more efficient uncertainty quantification.

% -------------------- TABLE: BLCS performance --------------------
\begin{table}[ht]
\centering
\caption{Comparative Performance of Survival Models on the BLCS Inference Set}
\label{tab:ESM-performance}
\begin{tabular}{lcccccccc}
\hline
Method & Mean C  & C-index SE & C-index AIL & Mean AUC   & AUC SE & AUC AIL \\
\hline
\textbf{ESM} & 0.7316  & 0.0018 & 0.0074 & 0.7987 & 0.0039 & 0.0171 \\
DeepSurv     & 0.7284  & 0.0019 & 0.0076 & 0.7928 & 0.0043 & 0.0186 \\
CoxPH        & 0.7223  & 0.0020 & 0.0080 & 0.7944 & 0.0050 & 0.0214 \\
RF           & 0.7221  & 0.0020& 0.0080 & 0.7928 & 0.0058 & 0.0247 \\
GBR          & 0.7209  & 0.0019 & 0.0076 & 0.7865 & 0.0043 & 0.0184 \\
\hline
\end{tabular}
\end{table}

% -------------------- FIGURE: performance plots --------------------
\begin{figure}[h!]
    \centering
    \begin{subfigure}[b]{0.48\textwidth}
        \centering
        \includegraphics[width=\textwidth]{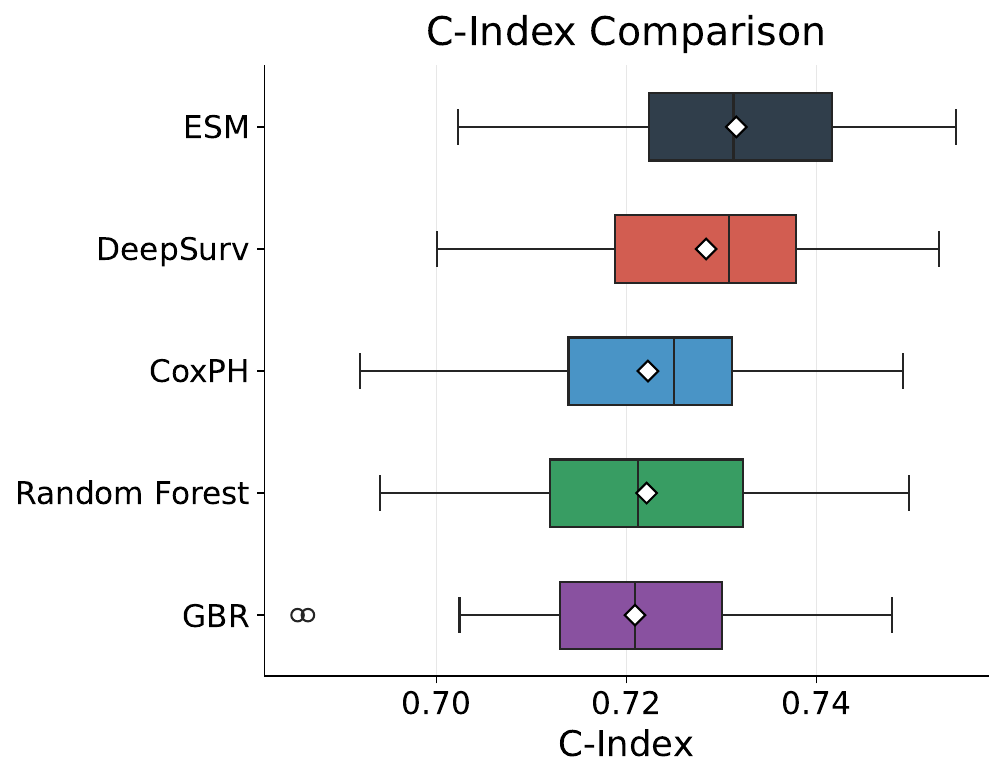}
        \caption{C-Index Comparison Boxplot}
        \label{fig:cindex_boxplot}
    \end{subfigure}
    \hfill
    \begin{subfigure}[b]{0.48\textwidth}
        \centering
        \includegraphics[width=\textwidth]{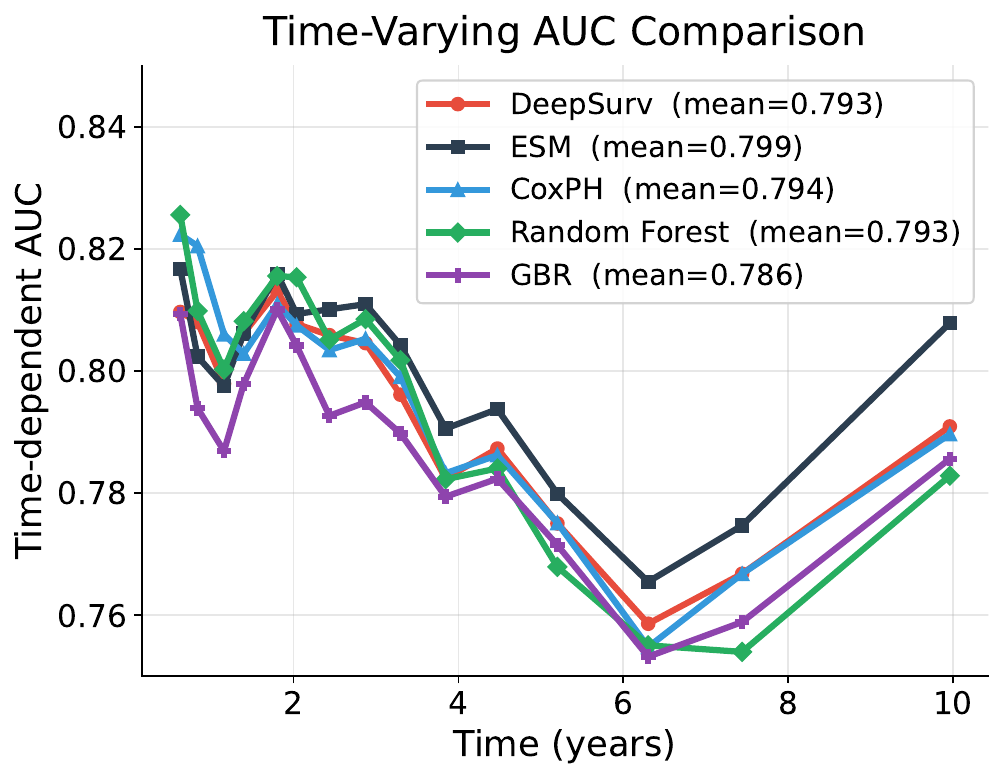}
        \caption{Time-Varying AUC Comparison}
        \label{fig:auc_timevarying}
    \end{subfigure}
    \caption{Comparative evaluation of model performance. The ensemble shows strong ranking stability and stable estimation accuracy over time.}
    \label{fig:performance_plots}
\end{figure}
\begin{figure}[htbp]
    \centering
    \begin{subfigure}[b]{0.48\textwidth}
        \centering
        \includegraphics[width=\textwidth]{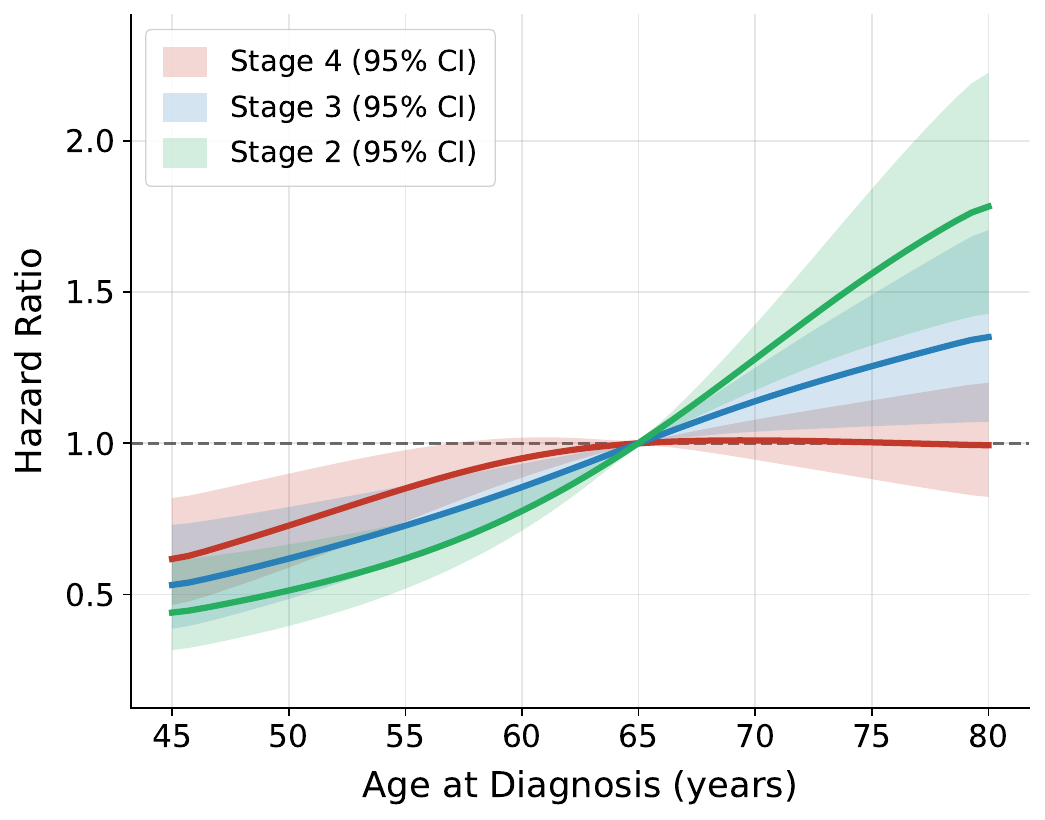}
        \caption{Age at Diagnosis (45 to 80 years)}
    \end{subfigure}
    \hfill
    \begin{subfigure}[b]{0.48\textwidth}
        \centering
        \includegraphics[width=\textwidth]{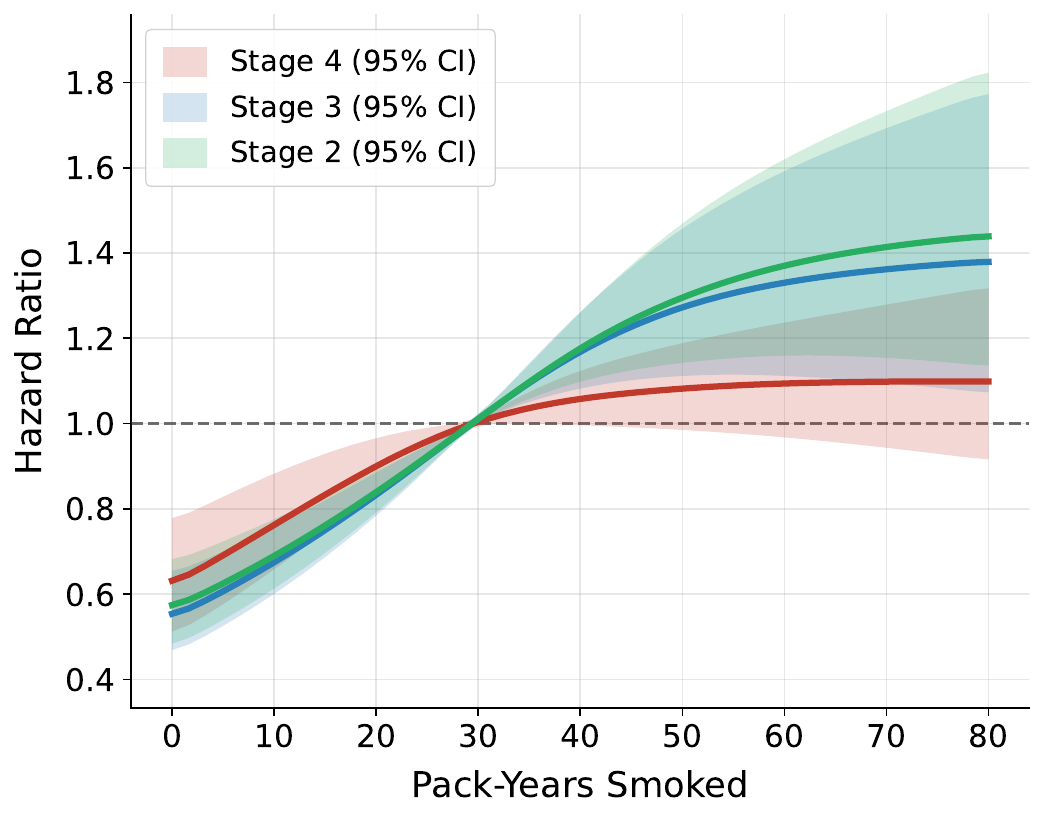}
        \caption{Pack-Years (0 to 80)}
    \end{subfigure}
    \vskip\baselineskip
    \begin{subfigure}[b]{0.48\textwidth}
        \centering
        \includegraphics[width=\textwidth]{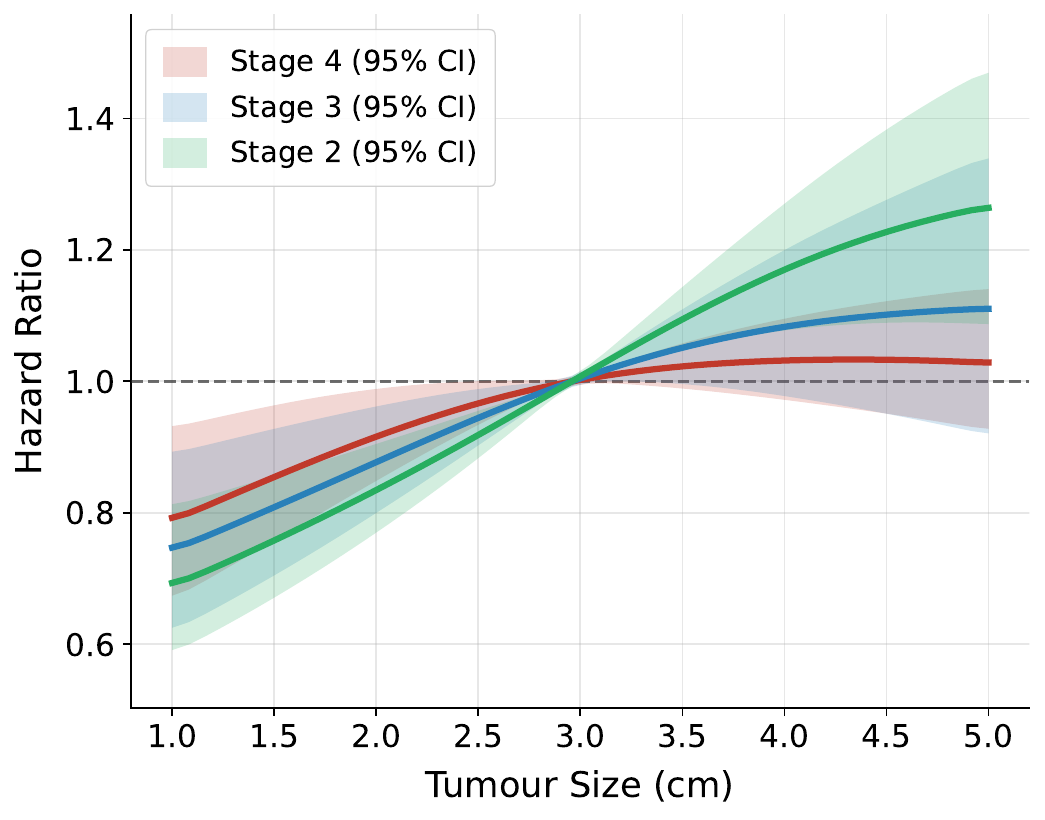}
        
        \caption{Tumor Size (2 to 6 cm)}
    \end{subfigure}
    \hfill
    \begin{subfigure}[b]{0.48\textwidth}
        \centering
        \includegraphics[width=\textwidth]{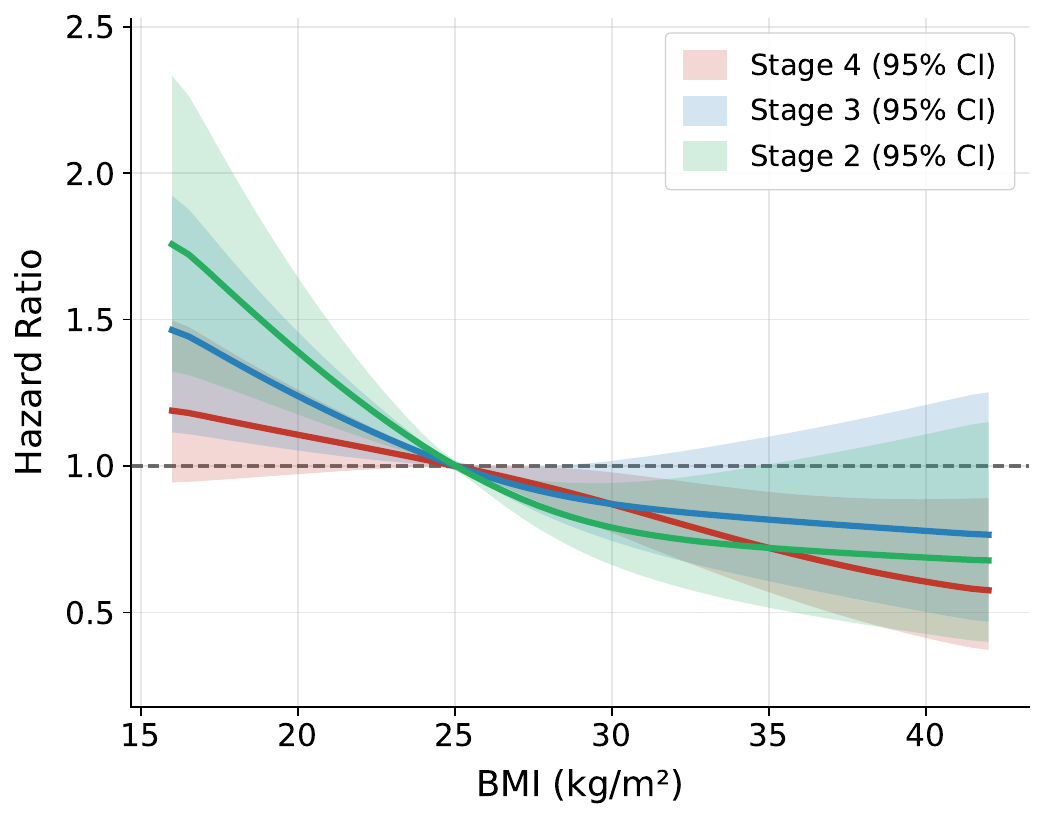}
        \caption{BMI (15 to 45 kg/m$^2$)}
    \end{subfigure}
    \caption{Stage-specific contrast HR studies. In each panel, curves correspond to Stages 2, 3, and 4, with shaded regions indicating IJ-based 95\% confidence bands.}
    \label{fig:hr_plots}
\end{figure}

 Stage-specific hazard ratios (HRs) as functions of patient characteristics are shown in Figure~\ref{fig:hr_plots}. Each curve is defined relative to a reference patient (male, received chemotherapy, age 65, BMI 26, 35 pack-years), with one covariate varied at a time and others fixed at their reference values.
Evaluation is performed over 50 evenly spaced values within clinically relevant ranges: age at diagnosis (45–80 years), tumor size (2–6 cm), smoking history (0–80 pack-years), and BMI (15–45 kg/m$^2$). The HR at each value compares the risk of death to that of the reference profile; values above 1 indicate increased risk, values below 1 indicate reduced risk, and 1 indicates no difference.
Uncertainty is quantified by pointwise 95\% confidence bands based on the infinitesimal jackknife (IJ) covariance estimator.
% -------------------- FIGURE: HR plots --------------------
 Confidence bands widen near the boundaries of the covariate ranges  with reduced data support in these regions. The stage-stratified contrast analysis characterizes variation in covariate effects across stages while providing valid uncertainty quantification via the IJ-based framework.

\section{Conclusion}
\label{sec:conc}
  We develop a unified theoretical and inferential framework for fully nonparametric Cox models with DNNs, linking optimization-based risk bounds to ensemble-based asymptotic inference. The resulting uncertainty quantification targets contrast functionals that are often more clinically relevant than absolute risk, such as hazard ratios comparing two patients who differ in a single feature (e.g., age 70 versus 50). Our methods yield stable variance estimation and reliable finite-sample coverage, addressing limitations of standard bootstrap methods in high-dimensional settings.

Future work follows naturally from the flexibility of the ensemble framework. First, we will extend inference from the risk function $g_0(\cdot)$ to the survival function $S(t\mid\mathbf{x}_*)$. Since survival probabilities are nonlinear functionals of the log-hazard, the Gaussian limits for $\hat g(\cdot)$ enable pointwise intervals and simultaneous bands via an infinite dimensional delta method, accounting for uncertainty in both the DNN and the Breslow-type baseline hazard estimator. 
Second, the framework extends naturally to treatment effect estimation in time-to-event settings. Building on ensemble-based estimation ideas \citep{Meng2025-zh}, we define the survival Conditional Average Treatment Effect (CATE) as
\[
\tau(t \mid \mathbf{x}_*) = S_{1}(t \mid \mathbf{x}_*) - S_{0}(t \mid \mathbf{x}_*),
\]
where $S_j(t \mid \mathbf{x}_*)$ denotes the conditional survival probability under treatment $j \in \{0,1\}$. Under standard unconfoundedness conditions, this can be estimated by applying ESM separately to the treated and control cohorts. Because the groups are independent, the variance of $\hat{\tau}(t\mid\mathbf{x}_*)$ is the sum of the group-specific IJ variances, yielding valid Wald-type confidence intervals. A full theoretical treatment of this causal extension remains for future work.

Finally, while our theory assumes controlled optimization error, further work is needed to understand how implicit regularization from practical training strategies, such as early stopping and dropout, interacts with the Cox loss and affects inference.

\bibliography{ref}
\bibliographystyle{ims}

\end{document}